\documentclass[11pt]{article} 
\usepackage[top=30truemm,bottom=30truemm,left=25truemm,right=25truemm]{geometry}
\usepackage{amsmath,amssymb}
\usepackage{ascmac}
\usepackage{graphicx}
\usepackage{verbatim} 
\usepackage{bm}
\usepackage{algorithm}
\usepackage{algorithmic}

\newcommand{\R}{\mathbb{R}}
\newcommand{\T}{\mathsf{T}}
\usepackage{authblk}
\usepackage{arydshln}
\usepackage{url}
\usepackage{caption}
\usepackage{lscape}
\usepackage{mathrsfs}
\usepackage[toc]{appendix}
\usepackage[T1]{fontenc}
\usepackage{lmodern}

\newcommand{\keywords}[1]{\par\addvspace\baselineskip\noindent\enspace\ignorespaces{\bf Keywords: }#1}

\title{Multiclass spectral feature scaling method \\for dimensionality reduction}
\author{Momo Matsuda\footnote{Corresponding author\\  Complete address of corresponding author: Momo Matsuda, Department of Computer Science, University of Tsukuba, 1-1-1 Tennodai, Tsukuba, Ibaraki, Japan; Tel: +81-29-853-6574; Fax: +81-29-853-6574; Email: \break matsuda.momo.ww@alumni.tsukuba.ac.jp}, Keiichi Morikuni, Akira Imakura, Xiucai Ye, Tetsuya Sakurai}
\affil{University of Tsukuba, Tsukuba, Japan}
\date{}

\begin{document}

\maketitle

\begin{abstract}
Irregular features disrupt the desired classification.
In this paper, we consider aggressively modifying scales of features in the original space according to the label information to form well-separated clusters in low-dimensional space.
The proposed method exploits spectral clustering to derive scaling factors that are used to modify the features.
Specifically, we reformulate the Laplacian eigenproblem of the spectral clustering as an eigenproblem of a linear matrix pencil whose eigenvector has the scaling factors.
Numerical experiments show that the proposed method outperforms well-established supervised dimensionality reduction methods for toy problems with more samples than features and real-world problems with more features than samples.
\end{abstract}

\section{Introduction}
Dimensionality reduction is a technique for reducing the number of variables of data samples and has been successfully applied in many fields to make machine learning algorithms faster and more accurate, including the pathological diagnoses of gene expression data~\cite{gene}, the analysis of chemical sensor data~\cite{sensor}, the community detection in social networks~\cite{network}, the analyses of neural spike sorting~\cite{spike}, and others~\cite{detection}.
Due to their dependence on label information, dimensionality reduction methods can be divided into supervised and unsupervised methods.
Typical unsupervised dimensionality reduction methods are the principal component analysis (PCA)~\cite{pca1,pca}, the classical multidimensional scaling (MDS)~\cite{mds}, the locality preserving projections (LPP)~\cite{lpp}, and the t-distributed stochastic neighbor embedding (t-SNE)~\cite{tsne}.
\par
To make use of prior knowledge on the labels, we focus on supervised dimensionality reduction methods.
Supervised dimensionality reduction methods map data samples into an optimal low-dimensional space for satisfactory classification while incorporating the label information.
One of the most popular supervised dimensionality reduction methods is the linear discriminant analysis (LDA)~\cite{lda}, which maximizes the between-class scatter and reduces the within-class scatter in a low-dimensional space.
However, LDA is based on the assumption that the input data obeys a Gaussian distribution and may fail to capture the underlying local data structure.
To overcome its drawback, variants of LDA have been proposed.
Bressan and Vitri\`a~\cite{nda} proposed a variant that preserves the local data structure by redefining the between-class scatter and within-class scatter matrices based on the $k$-nearest neighbors of each data sample.
Weinberger and Saul~\cite{ML} considered penalizing large distances among data samples with the same label and the $k$-nearest neighbors and small distances among data samples with different labels according to a cost function.
Sugiyama proposed the local Fisher discriminant analysis (LFDA)~\cite{lfda2}, which maximizes the between-class separability and preserves the within-class local structure, incorporating LPP for supervised dimensionality reduction.
Cai et al.~\cite{lsda} proposed another variant, which captures the local geometrical data structure to map $k$-nearest neighbors in the same classes and those in different classes to a low-dimensional space as close and as distant as possible, respectively.
Instead of preserving the local data structure based on the distances among data samples in the original data space, Li et al.~\cite{lada} introduced a weight between a pair of samples in the same class and exploited the local manifold structure of data samples in a low-dimensional space.
\par
In addition to capturing the local structure, as in LDA and its variants, we consider aggressively modifying the scales of the features in the original space before mapping the data samples to a low-dimensional space.
This results in reducing the effect of features that interfere in the desirable classification as well as enhancing the separations of different clusters.
For these purposes, we derive quantities called feature scaling factors, which are used to scale features based on spectral clustering (SC) for dimensionality reduction~\cite{scs}.
SC projects given data samples to a low-dimensional space formed by eigenvectors of a similarity or Laplacian matrix.
In~\cite{sc1,sc2}, it was reported that SC is effective in detecting cluster structures, and in~\cite{sc3}, it was reported that SC preserves the local data structure for feature selection.
These reports motivated us to also use SC to learn local cluster structures.
\par
Our purpose is to scale features of given data samples and form (well-separated) clusters in a lower dimensional space.
Previous spectral feature scaling methods aimed at binary classification~\cite{scs}: in this paper, we extend the scope of classification problems to handle multiclassing.
Multiclass classification problems with more than two classes have typically been solved by combining independently produced binary classification problems~\cite{svm-multi}.
We take an approach different from~\cite{svm-multi} to enable a multiclass classification.
The contributions of this study can be summarized as follows:
\begin{description}
\item[{\it Extension to multiclass classification}] \mbox{}\\
 We extend the scope of problems that the spectral feature scaling method can handle to multiclass classification.
 Previous formulations allow only binary classification and require repeated applications of the method for multiclass classification.
\item[{\it Automatic parameter tuning}] \mbox{}\\
We propose an automatic tuning technique to determine the values of parameters that specify entries of the (Fielder) eigenvectors according to the labels of training data samples.
The extension of the spectral feature scaling method to multiclass classification naively increases the number of hyperparameters and the computational cost.
Our technique can save this cost in the multiclass case.
\end{description}
The remainder of this paper is organized as follows.
In Section~2, we present preliminaries.
In Section~3, we present the proposed spectral feature scaling method.
In Section~4, we describe the difference between the proposed method and previous methods.
In Section~5, we present experimental results comparing the proposed method with previous methods.
In Section~6, we conclude the paper.
Throughout the paper, we use italic bold characters to represent column vectors.

\section{Preliminaries}
Conventional spectral clustering nonlinearly reduces the dimensionality of data samples but may not form the desired data sample clusters due to irregularity of the data features.
To improve the classification of reduced data samples, we may modify the scales of features.
Simple feature scaling techniques, such as centering, standardization, and normalization, serve as preprocessing for the support vector machine and $k$-means algorithm.
This motivated us to propose supervised feature scaling that enhances the separation of reduced data samples based on prior knowledge from training data.
\par
The spectral clustering partitions a weighted undirected graph in which the edge weight $w_{i,j}$ between nodes $i$ and $j$ is defined according to the similarity between data samples $\bm{x}_i$ and $\bm{x}_j$, where $i, j = 1,2,\ldots,n$ and $n$ is the number of data samples.
We use the locally scaled Gaussian kernel function~\cite{sigma} to define the similarity between data samples $\bm{x}_i$ and $\bm{x}_j$ as follows:
\begin{eqnarray}
w_{i,j} = \exp\left(- \frac{\left( \bm{x}_i - \bm{x}_j\right)^{\T} S \left(\bm{x}_i - \bm{x}_j\right)}{\sigma^{\prime}_i\sigma^{\prime}_j}\right),
\label{eq:wij}
\end{eqnarray}
where $S = {\rm diag}\left(s_1, s_2, \ldots, s_m\right)\in\R^{m\times m}$, and $\sigma^{\prime}_i$ denotes the local scaling of data samples around $\bm{x}_i$.
Thus, we convert a given set of data samples to a graph according to similarities or distances among data samples.
A similarity graph is given by a model of the local relationship between data samples.
We set $\sigma^{\prime}_i = \left(\bm{x}_i - \bm{x}_i^{(k)}\right)^{\T} S \left(\bm{x}_i - \bm{x}_i^{(k)}\right)$, 
where $\bm{x}_i^{(k)}$ is the $k$th nearest neighbor of $\bm{x}_i$.
Zelnik-Manor and Perona~\cite{sigma} recommended the value $k=7$ for $S=I$ in spectral clustering in practice, and we adopted this choice.
\par
Spectral clustering partitions a graph into subgraphs that are contradictory to each other and forms clusters from the corresponding data samples.
Let $W_s=\left(w_{i,j}\right)\in\R^{n\times n}$ be the similarity matrix,
\begin{eqnarray}  
  D_s=\mathrm{diag}\left(d_1,d_2,\ldots,d_n\right)\in\R^{n\times n} \quad \mbox{ with } \quad d_i = \sum_{j=1}^n w_{i,j}, 
      \label{eq:WandD}
\end{eqnarray}
$\bm{e}$ be a vector of all ones, and $t_i \in \lbrace \pm 1 \rbrace$ be an indicator such that $t_i = 1$ when data sample $i$ belongs to a cluster and $t_i=-1$ when data sample $i$ belongs to another cluster. 
Then, the graph partitioning is given by the constrained minimization problem of the normalized cut (Ncut) function~\cite{sc}:
\begin{eqnarray}
  &\min_{\bm{v}} \displaystyle\frac{\bm{v}^{\mathsf{T}} (D_s - W_s) \bm{v}}{\bm{v}^{\mathsf{T}}D_s\bm{v}}, \nonumber \\ 
  &\mbox{subject to} \quad \bm{e}^{\mathsf{T}}D_s^{(p)}\bm{v} = 0, \quad v_i \in \{ 1, -b \}, \  b=\frac{\sum_{i:t_i > 0}d_i}{\sum_{i:t_i<0}d_i}.
  \label{eq:j}
\end{eqnarray}
The entry $v_i \in \lbrace 1, -b \rbrace$ takes two discrete values to devide the set of data samples into two clusters.
By the continuous relaxation $v_i \in \lbrace 1, -b \rbrace$ to $v_i \in \R$, (\ref{eq:j}) is reduced to finding the Fiedler vector $\bm{v}\in\R^n\setminus\{ \bm{0}\}$~\cite{fiedler} corresponding to the minimum eigenvalue of the constrained generalized eigenvalue problem
\begin{equation}
  L_s\bm{v} = \lambda_s D_s \bm{v}, \quad \lambda_s \in\R, \quad \mbox{subject to} \quad \bm{e}^{\mathsf{T}}D_s\bm{v} = 0,
  \label{eq:lap}
\end{equation}
if the minimum eigenvalue has multiplicity one, where $L_s = D_s - W_s$ is the Laplacian matrix.
The entries of the Fiedler vector are referred to as the coordinates of the data samples in the reduced space.
In practice, we use more eigenvectors corresponding to the minimum eigenvalues of (\ref{eq:lap}) to form a larger space for accurate classification.
When clustering well-separated data samples, we will find hyperplanes that separate the data samples with reduced dimensions formed by a few eigenvectors in (\ref{eq:lap}).
Here, if $S=I$ in (\ref{eq:wij}), i.e., $S$ is the identity matrix, (\ref{eq:lap}) is reduced to a conventional spectral clustering method with relaxed Ncut~\cite{sc}.
If $S \neq I$, (\ref{eq:lap}) is a generalization of the method~\cite{sc} and we can calculate the diagonal entries $s_i \ (i =1,2,\ldots,m)$ of $S$, which are called the scaling factors, for each data feature for binary classification problems~\cite{scs}.

\section{Proposed method}
In this section, we formulate the spectral feature scaling method (SFS) for dimensionality reduction.
The method was originally proposed for binary clustering and classification, but it is here extended to multiclass classification.
The proposed method does not use typical means of extension to multiclass classification, such as multiclass support vector machines~\cite{svm-multi}.
The new formulation of SFS computes $r$ scaling factors for each feature and merges them into a single scaling factor.
Here, the parameter $r$ is the number of combinations that divide the set of data samples into two classes: we will discuss the selection of $r$ in Section~3.3.
\par
If the labels of the training data samples are known, we use discrete variable $v_i \in \lbrace 1, -b \rbrace$ in~(\ref{eq:j}) to prescribe the entries of the $r$ eigenvectors $\bm{v}^{(p)}\in\R^n\setminus \lbrace \bm{0} \rbrace \ (p=1,2,\ldots,r)$ of the Laplacian eigenvalue problem 
\begin{equation}
  L_s^{(p)} \bm{v}^{(p)} = \lambda_s^{(p)} D_s^{(p)} \bm{v}^{(p)}, \quad \bm{e}^{\T}D_s\bm{v}^{(p)} = 0, \quad \lambda_s^{(p)} \in \R,
  \label{eq:LS}
\end{equation}
where $\bm{v}^{(p)} \in \lbrace 1, -b^{(p)}\rbrace$, $L_s^{(p)} = D_s^{(p)} - W_s^{(p)}$, $D_s^{(p)} = \mathrm{diag}\left(d_1^{(p)},d_2^{(p)},\ldots,d_n^{(p)}\right)$, and $W_s^{(p)} = \left(w_{i,j}^{(p)}\right) \in\R^{n\times n}$ depend on the feature scaling factors $s_i^{(p)}\in\R, i = 1,2, \ldots, m$, and $b^{(p)}$ is a parameter: we will discuss how to estimate the values of $b^{(p)}$ in Section~3.3.
The entries $w_{i,j}^{(p)}$ and $d_i^{(p)}$ are analogously defined as (\ref{eq:WandD}).
We form the scaling matrix $S$ from $s_i^{(p)}$ and will discuss the detail in Section~3.2.
Then, we apply $S^{1/2}$ to data $[X^\T,\ T^\T]^\T$, which consists of the training data $X=[\bm{x}_1,\bm{x}_2,\ldots,\bm{x}_n]^\T\in\R^{n\times m}$ and the test data $Y\in\R^{N\times m}$ as follows:
\begin{eqnarray}
Z = \left[
\begin{array}{c}
X\\
Y
\end{array}\right] S^{1/2}.
\label{eq:6}
\end{eqnarray}
With the feature scaling $S$, we enhance the separation of the reduced data samples as desired in the low-dimensional space formed by spectral clustering.
This will improve the classification results.
Finally, we classify the scaled data $Z\in\R^{(N+n) \times m}$ using $\ell$ eigenvectors corresponding to the positive minimum eigenvalue of the Laplacian eigenvalue problem
\begin{equation*}
  L' \bm{u} = \lambda' D' \bm{u}, \quad \bm{e}^{\T}\!D'\bm{u} = 0, \quad \bm{u}\in\mathbb{R}^N\setminus\lbrace\bm{0}\rbrace, \quad \lambda' \in \mathbb{R},
\end{equation*}
where $L'=D'-W'\in\R^{N \times N}, W' = \left( w'_{i,j} \right) \in \R^{N \times N},$
\begin{equation}
  w'_{i,j} = \exp\left(-\frac{{\| \bm{z}_i - \bm{z}_j \|_2}^2 }{\sigma_i\sigma_j}\right),\quad  i, j = 1, 2 \ldots, N,
  \label{eq:w}
\end{equation}
$\bm{z}_i\in\R^m$ is the $i$th column of $Z^\T$, $d'_i=\sum_{j=1}^n w'_{i,j}, D'=\mathrm{diag} \left(d'_1, d'_2, \ldots, d'_N\right)$, and $\sigma_i = \|\bm{z}_i - \bm{z}_i^{(k)}\|_2$.
The values of parameters $\ell$ and $r$ can be determined by cross-validation.

\subsection{Formulation}
Now, we reformulate (\ref{eq:LS}) as another eigenvalue problem to extract the scaling factors $\sqrt{s_i^{(p)}} \ (i = 1,2,\ldots,m)$ as an eigenvector. Denote the scaling matrix by ${S^{(p)}}^{1/2} = \mathrm{diag}(\sqrt{s_1^{(p)}}, \sqrt{s_2^{(p)}}, \ldots, \sqrt{s_m^{(p)}})\in\R^{m \times m}$ and the $(i,j)$ entry of the similarity matrix $W_s^{(p)}$ of scaled data $X{S^{(p)}}^{1/2}$ by
\begin{eqnarray}
  w_{i,j}^{(p)} = \left\{
  \begin{array}{lr}
    \quad 1 - \frac{\left( \bm{x}_i - \bm{x}_j\right)^{\T} S^{(p)} \left(\bm{x}_i - \bm{x}_j\right)}{\sigma_i\sigma_j}\\
    = 1 - \frac{\bm{s}^{(p)\T} \bm{x}_{i,j}}{\sigma_i\sigma_j}
    \simeq \exp\left(- \frac{\left( \bm{x}_i - \bm{x}_j\right)^{\T} S^{(p)} \left(\bm{x}_i - \bm{x}_j\right)}{\sigma^{\prime}_i\sigma^{\prime}_j}\right), & \quad i \neq j, \\
    \quad 0, & \quad i = j,
  \end{array}\right. \label{eq:ws}
\end{eqnarray}
where $\sigma_i = \|\bm{x}_i - \bm{x}_i^{(k)}\|_2$, $\bm{s}^{(p)} = \left[ s_1^{(p)}, s_2^{(p)}, \ldots, s_m^{(p)} \right]^{\T} \in \R^{m}$, and the $k$th entry of $\bm{x}_{i,j}\in\R^m$ is $(x_{i,k} - x_{j,k})^2$~\cite{wij}.
Here, we used the first-order approximation of the exponential function $\exp\left(-x\right)\simeq~1-x$ \ for \ $0 < x \ll 1$. Then, the $i$th row of $W_s^{(p)}$ is 
\begin{equation*}
  \bm{w}_i^{(p)\T} = \left[1,\ldots,1,0,1,\ldots,1\right] -\bm{s}^{(p)\T}\left[ \frac{\bm{x}_{i,1}}{\sigma_i\sigma_1}, \frac{\bm{x}_{i,2}}{\sigma_i\sigma_2},\ldots,\frac{\bm{x}_{i,n}}{\sigma_i\sigma_n}\right] = \tilde{\bm{e}}_i^\T - \bm{s}^{(p)\T} X_i,
\end{equation*}
where $\tilde{\bm{e}}_i$ is an $n$-dimensional vector with zero in the $i$th entry and ones in all other entries, and $X_i = \left[\frac{\bm{x}_{i,1}}{\sigma_i\sigma_1},\frac{\bm{x}_{i,2}}{\sigma_i\sigma_2},\ldots,\frac{\bm{x}_{i,n}}{\sigma_i\sigma_n}\right] \in\R^{m\times n}$. Hence, we have
\begin{eqnarray*}
  W_s^{(p)}\bm{v}^{(p)} = \left[\tilde{\bm{e}}_1^\T,\tilde{\bm{e}}_2^\T,\ldots,\tilde{\bm{e}}_n^\T\right]\bm{v}^{(p)} - \left[X_1\bm{v}^{(p)},X_2\bm{v}^{(p)},\ldots,X_n\bm{v}^{(p)}\right]^\T \bm{s}^{(p)}.
\end{eqnarray*}
Let $\hat{\bm{x}}_i = \sum_{j=1}^n \frac{\bm{x}_{i,j}}{\sigma_i\sigma_j}$. Then, the $i$th diagonal entry of $D_s^{(p)}$ is
\begin{eqnarray*}
  d_i^{(p)} = \sum_{j=1}^n w_{i,j}^{(s)} = \left(n-1\right) - \bm{s}^{(p)\T}\hat{\bm{x}}_i.
\end{eqnarray*}
Hence, denoting the eigenvector by $\bm{v}^{(p)} = \left[v_1^{(p)},v_2^{(p)},\ldots,v_n^{(p)}\right]^\T$, we have
\begin{eqnarray*}
  D_s^{(p)}\bm{v}^{(p)} = \left(n-1\right)\bm{v}^{(p)} - \left[v_1^{(p)}\hat{\bm{x}}_1,v_2^{(p)}\hat{\bm{x}}_2,\ldots,v_n^{(p)}\hat{\bm{x}}_n\right]^\T \bm{s}^{(p)}.
\end{eqnarray*}
Thus, (\ref{eq:LS}) can be written as
\begin{eqnarray}
 && L_s^{(p)} \bm{v}^{(p)} = \lambda_s^{(p)} D_s^{(p)} \bm{v}^{(p)} \nonumber \\
  &\Leftrightarrow& W_s^{(p)}\bm{v}^{(p)} = \left(1-\lambda_s^{(p)}\right)D_s^{(p)}\bm{v}^{(p)} \nonumber \\
                                                      &\Leftrightarrow& 
                                                                        \begin{bmatrix}
                                                                          A^{(p)} & \bm{\alpha}^{(p)}
                                                                        \end{bmatrix}\begin{bmatrix}{}
                                                                                      \bm{s}^{(p)}\\
                                                                                      -1
                                                                                    \end{bmatrix} = \mu^{(p)}
  \begin{bmatrix}
    B^{(p)} & \bm{\beta}^{(p)}
  \end{bmatrix}\begin{bmatrix}
                \bm{s}^{(p)}\\
                -1
              \end{bmatrix},                  
              \label{eq:9}                          
\end{eqnarray}
where $\mu^{(p)} = 1-\lambda_s^{(p)}$,
\begin{eqnarray}
  A^{(p)}=\left[\begin{array}{c}
                  \left(X_1\bm{v}^{(p)}\right)^\T\\
                  \left(X_2\bm{v}^{(p)}\right)^\T\\
                  \vdots\\
                  \left(X_n\bm{v}^{(p)}\right)^\T
                \end{array}\right]\in\R^{n\times m}, \quad B^{(p)}=\left[
  \begin{array}{c}
    v_1^{(p)}\hat{\bm{x}}_1^{\T}\\
    v_2^{(p)}\hat{\bm{x}}_2^{\T}\\ 
    \vdots\\
    v_n^{(p)}\hat{\bm{x}}_n^{\T}
  \end{array}\right]\in\R^{n\times m}, \label{eq:AB}\\
  \bm{\alpha}^{(p)} = \left[\tilde{\bm{e}}_1, \tilde{\bm{e}}_2, \ldots, \tilde{\bm{e}}_n\right]^\T\bm{v}^{(p)}\in\R^n, \quad \bm{\beta}^{(p)} = (n-1)\bm{v}^{(p)}\in\R^n. 
\label{eq:alphabeta}
\end{eqnarray}
Furthermore, the feature scaling factors have the constraint
\begin{eqnarray}
  && \bm{e}^\T D_s^{(p)}\bm{v}^{(p)} = \sum_{i=1}^n\left(\left(n-1\right)v_i^{(p)} - v_i^{(p)}\bm{s}^{(p)}\hat{\bm{x}}_i\right) = 0 \nonumber \\
  && \Leftrightarrow \left(\sum_{i=1}^nv_i^{(p)}\hat{\bm{x}}_i^\T\right)\bm{s}^{(p)} - \left(n-1\right)\sum_{i=1}^nv_i^{(p)} = 0.
  \label{eq:12}
\end{eqnarray}
Combining (\ref{eq:9})--(\ref{eq:12}), we can obtain the generalized eigenvalue problem of a linear matrix pencil
\begin{eqnarray}
  \mathcal{A}^{(p)}
  \begin{bmatrix}
    \bm{s}^{(p)}\\
    -1
  \end{bmatrix} = \mu^{(p)} \mathcal{B}^{(p)}
  \begin{bmatrix}
    \bm{s}^{(p)}\\
    -1
  \end{bmatrix}, \quad \bm{s}^{(p)}\in\R^m, \quad \mu^{(p)}\in\R,
  \label{eq:rec2}
\end{eqnarray}
where
\begin{eqnarray}
  \mathcal{A}^{(p)} = 
  \begin{bmatrix}
    A^{(p)} & \bm{\alpha}^{(p)}\\
    \bm{\gamma}^{(p)\T} & \rho^{(p)}
  \end{bmatrix}\in\R^{(n+1)\times (m+1)}, \label{eq:A} \\
  \mathcal{B}^{(p)} = 
  \begin{bmatrix}
    B^{(p)} & \bm{\beta}^{(p)}\\
    \bm{0}^\T & 0
  \end{bmatrix}\in\R^{(n+1)\times (m+1)}, \label{eq:AandB}\\
  \bm{\gamma}^{(p)} = \sum_{i=1}^nv_i^{(p)}\hat{\bm{x}}_i\in\R^m, \quad \rho^{(p)} = (n-1)\sum_{i=1}^nv_i^{(p)}\in\R, \label{eq:gammarho}
\end{eqnarray}
and we solve (\ref{eq:rec2}) for the scaling factors $\bm{s}^{(p)}$ for each $p = 1,2, \ldots, r$.
If the $r$ eigenproblems (\ref{eq:rec2}) have a common eigenpair, they can be combined as
\begin{eqnarray}
  \begin{bmatrix}
    \mathcal{A}^{(1)}\\
    \mathcal{A}^{(2)}\\
    \vdots\\
    \mathcal{A}^{(r)}\\
  \end{bmatrix}
  \begin{bmatrix}
    \bm{s}\\
    -1
  \end{bmatrix} = \mu_s
  \begin{bmatrix}
    \mathcal{B}^{(1)}\\
    \mathcal{B}^{(2)}\\
    \vdots\\
    \mathcal{B}^{(r)}\\
  \end{bmatrix}
  \begin{bmatrix}
    \bm{s}\\
    -1
  \end{bmatrix}, \quad \bm{s} \in\R^m, \quad \mu_s\in\R,
  \label{eq:LS2}
\end{eqnarray}
where $\mu_s=1-\lambda_s$.
The existence of an eigenvalue $\mu^{(p)}$ of (\ref{eq:rec2}) is assured by the equality\linebreak
$\left[\bm{e}^\T\ 0\right]\left(\mathcal{A}-\mathcal{B}\right)^\T=\bm{0}^\T$.
Because the Fiedler vector is associated with the smallest eigenvalue, we adopt the eigenvector of (\ref{eq:rec2}) and (\ref{eq:LS2}) associated with the eigenvalues $\mu^{(p)} = 1 - \lambda_s^{(p)}$ and $\mu_s$ closest to but less than one as a candidate feature scaling factors, respectively.
Here, the $r$ eigenproblem (\ref{eq:LS2}) may not have the same eigenpair, so we find an eigenpair $\left(\lambda, \ [\bm{s}^\T, -1]^\T\right)$ of (\ref{eq:A}) and (\ref{eq:AandB}) as close as possible in the $r$ eigenproblems.
To solve (\ref{eq:rec2}) or (\ref{eq:LS2}), we can use the contour integral method~\cite{ss} for $n<m$; otherwise, we can also use the minimal perturbation approach~\cite{svd}.

\subsection{Integrating candidates of feature scaling factors}
Here, we describe how to determine the scaling factors from the candidates in the multiclass case.
The scaling factors $\sqrt{s_i^{(p)}}$ given by solving (\ref{eq:rec2}) are used to form a single feature scaling matrix $S^{1/2}={\rm diag}\left(s_1,s_2,\ldots,s_m\right)^{1/2}$.
We have the following five approaches to computing the diagonal elements $s_i^{1/2} \ (i = 1,2,\ldots,m)$ of $S^{1/2}$:
\begin{itemize}
\item Principal component analysis (PCA)~\cite{pca1,pca}\\
We can obtain $s_i^{1/2}$ by using the principal component, i.e., the eigenvector $\bm{\phi}\in\R^r$ associated with the maximum eigenvalue $\tau\in\R$ of the eigenproblem
\begin{equation*}
M \bm{\phi} = \tau \bm{\phi}, \quad M = \sum_{i=1}^m \left(\bm{\hat{s}}_i-\hat{\bm{\mu}} \right)\left(\bm{\hat{s}}_i-\hat{\bm{\mu}} \right)^\T,
\end{equation*}
where $\bm{\hat{s}}_i^\T \in\R^r$ is $i$th row of the matrix
\begin{eqnarray*}
\begin{bmatrix}
\sqrt{s_1^{(1)}} & \sqrt{s_1^{(2)}} & \ldots & \sqrt{s_1^{(r)}}\\
\sqrt{s_2^{(1)}} & \sqrt{s_2^{(2)}} & \ldots & \sqrt{s_2^{(r)}}\\
\vdots & \vdots & \ddots & \vdots\\
\sqrt{s_m^{(1)}} & \sqrt{s_m^{(2)}} & \ldots & \sqrt{s_m^{(r)}}\\
\end{bmatrix}\in\R^{m\times r},
\end{eqnarray*}
and $\hat{\bm{\mu}} = \frac{1}{m}\sum_{i=1}^m\bm{\hat{s}}_i$.
Thus, $s_i^{1/2} = \bm{\hat{s}}_i^\T\bm{\phi}$ is obtained.
\end{itemize}

\begin{itemize}
\item Arithmetic mean $s_i^{1/2} = \frac{1}{r}\sum_{p=1}^r \sqrt{s_i^{(p)}}$.
\item Geometric mean $s_i^{1/2} = \left(\prod_{p=1}^r \sqrt{s_i^{(p)}}\right)^{1/r}$.
\item Root mean square $s_i^{1/2} = \sqrt{\frac{1}{r}\sum_{p=1}^r s_i^{(p)}}$.
\item Harmonic mean $s_i^{1/2}=\left.r \middle/ \displaystyle \sum_{p=1}^r\left(1 \slash \sqrt{s_i^{(p)}}\right)\right.$.
\end{itemize}
We will compare the classification accuracy of each approach for integrating scaling factors in the numerical experiments.

\subsection{Number of prescribed eigenvectors and prescription of the entries}
Consider determining a reasonable number $r$ of prescribed eigenvectors $\bm{v}^{(p)}$ to prevent solving many eigenvalue problems (\ref{eq:rec2}) or a large eigenvalue problem (\ref{eq:LS2}).
Let $K$ be the number of classes.
In particular, if $K=2$, we prescribe only one eigenvector $\bm{v}^{(p)}, p = 1, r = 1$ in (\ref{eq:LS}), as in~\cite{scs}.
This is because one eigenvector is sufficient for binary classification if the clusters are well-separated~\cite{theory}.
If $K > 2$, we split the set of training data samples into $r$ combinations of binary classes with two options: take $r=K$ or take at least $r = \lceil \log_2K\rceil$ and determine the entries of the eigenvector $\bm{v}^{(p)}$ as
\begin{eqnarray}
  v_i^{(p)} = \left\{
  \begin{array}{cl}
    1 & \mbox{if sample } i \mbox{ is in class } c, \quad c=1,2,\ldots,K, \\
    -b^{(p)} & \mbox{otherwise}
  \end{array}\right.
               \label{eq:element}
\end{eqnarray}
for $p = 1,2,\dots,r$ according to the label information of the training data, where $b^{(p)}$ is a parameter we discuss how to determine later in this section.
Remark that the $r$-dimensional indicator vector $[v_i^{(1)}, v_i^{(2)},\ldots,v_i^{(r)}]$ for sample $i$ is uniquely chosen to a label.
This is a solution of discrete optimization for a binary clustering (\ref{eq:j}) reduced from multiclass classification.
That is, one cluster consists of data samples belonging to several classes, and the other clusters consist of data samples belonging to other classes.
The classification accuracy varies depending on the combination of such binary classifications.
By setting $r = K$, the computational cost for solving (\ref{eq:rec2}) or (\ref{eq:LS2}) is large, while the classification accuracy tends to be less sensitive to the choice of combinations.
When setting $r$ as small, the classification accuracy tends to be sensitive and the computational cost for solving (\ref{eq:rec2}) or (\ref{eq:LS2}) is small.
Hence, there is a trade-off between the cost and sensitivity of classification accuracy for these approaches.
\par
In the previous SFS with $r = 1$ in (\ref{eq:element}), we empirically determined the value of $b$~\cite{scs}.
In this study, we found that the values of $b^{(p)}$ in (\ref{eq:element}) determined by
\begin{eqnarray}
  b^{(p)} = \frac{\sum_{i:t_i>0}d_i^{(p)}}{\sum_{i:t_i<0}d_i^{(p)}}, \quad p=1,2,\ldots,r
  \label{eq:b}
\end{eqnarray}
according to the label information of training data $X$ give favorable results, where $t_i = 1$ indicates that data sample $i$ belongs to a cluster, and $t_i=-1$ indicates that data sample $i$ belongs to another cluster.
The choice of $t_i=1$ or $t_i=-1$ is flexible in (\ref{eq:b}) and can be fixed by cross-validation in practice (see Section 5).
Note that the ideal Fiedler vector does not necessarily take such an entry and is not necessarily unique.
\par\bigskip
We summarize the procedures of the proposed method in Algorithm~\ref{alg2}.

\begin{algorithm}[!h]             
  \caption{Supervised dimensionality reduction method with spectral features scaling}        
  \label{alg2}                          
  \begin{algorithmic}[1]
    \REQUIRE Training data $X \in \R^{n \times m}$ with $n$ samples and $m$ features, test data $Y \in \R^{N \times m}$, number of reduced dimensions $\ell$.
    \ENSURE Low-dimensional data samples $ U = [\bm{u}_1, \bm{u}_2, \dots, \bm{u}_{\ell}]^\T\in\R^{\ell \times (N+n)}$.
    \STATE Decide the number of eigenvectors to prescribe $r$.
    \STATE Compute the parameter $b^{(p)}$ using $X$ and set the eigenvectors $\bm{v}^{(p)}, \ (p = 1,2,\ldots,r)$.
    \STATE Compute $A^{(p)}, B^{(p)}, \bm{\alpha}^{(p)}, \bm{\beta}^{(p)}, \bm{\gamma}^{(p)}$, and $\rho^{(p)}$ using (\ref{eq:AB}), (\ref{eq:alphabeta}), and (\ref{eq:gammarho}).
    \STATE Solve (\ref{eq:rec2}) for an eigenvector $\bm{s}^{(p)} $.
    \STATE Compute the scaling matrix $S^{1/2}$ from the scaling factors $\bm{s}^{(p)}$ (Section~3.3).
    \STATE Compute the scaled data matrix $Z$ (\ref{eq:6}).
    \STATE Compute the similarity matrix $W'\in\R^{(N+n) \times (N+n)}$ (\ref{eq:w})  and $D'$.
    \STATE Compute the Laplacian matrix $L' = D' - W'$.
    \STATE Solve $L'\bm{u} = \lambda' D' \bm{u}$ for the eigenvectors $\bm{u}_1,\bm{u}_2 \ldots \bm{u}_\ell$ corresponding to the $\ell$ smallest nonzero eigenvalues.
  \end{algorithmic}
\end{algorithm}

\section{Related methods}
In this section, we discuss the difference between the proposed method and previous methods.
Many supervised dimensionality reduction methods, such as LDA and its variants, utilize knowledge from labels and map given data samples to an optimal low-dimensional space.
The objective of these methods is to find a transformation matrix $T$ to map given data samples $X$ to low-dimensional data samples $U$
\begin{equation*}
  U = TX ^\T,
\end{equation*}
whereas the objective of our method is to generate a scaling matrix $S^{1/2}$ and modify the scales of the features of given data samples $X$ as $XS^{1/2}$ and then map them to low-dimensional data samples $U^\prime$ using a transformation matrix $T^\prime$:
\begin{equation*}
  U^\prime = T^\prime \phi\left(S^{1/2}X^\T\right),
\end{equation*}
where $\phi:\R^{m\times n} \rightarrow \R^{m\times n}$ is a transformation with a nonlinear kernel function.
The methods in~\cite{lada,lada2} modify the weights among data samples, whereas the proposed method modifies the scales of features in the original space and uses the same feature scaling factors for all data samples.

\section{Numerical experiments}
We compare the proposed method (SFS) with previous supervised dimensionality reduction methods in terms of accuracy by numerical experiments on artificial data and practical datasets and show that the proposed method outperforms previous methods in some cases, and is more robust than the previous methods.
The previous methods were kernel versions of LDA (KDA)~\cite{kda,kda2} and LFDA (KLFDA)\cite{lfda} and LADA~\cite{lada}.
The performance measures used for comparisons were overall accuracy (OA), average accuracy (AA), and normalized mutual information (NMI)~[\%]~\cite{nmi}.
Here, OA and AA are defined as
\begin{eqnarray*}
{\rm OA} = \frac{\sum_{i=1}^K \hat{n}_i}{\sum_{i=1}^K n_i}, \quad {\rm AA} = \frac{1}{K}\sum_{i=1}^K \frac{\hat{n}_i}{n_i},
\end{eqnarray*}
where $n_i$ is the number of samples in class $i$, $\hat{n}_i$ is the number of samples classified by a method to class $i$, and $K$ is the number of classes.
We performed five-fold cross-validation and show the mean and standard deviation of each performance measure.
The reduced dimension $\ell$ was chosen by four-fold cross-validation regarding OA between one to the number of features when the number of features is smaller than the number of training data samples and between one to the number of training data samples, otherwise.
In SFS, we used the seven nearest neighbors to sparsify the matrix $W'$.
We chose optional values of the parameters $\lambda$ and $\gamma$ of LADA~\cite{lada2} among $1,10^{\pm1},10^{\pm2}$, and $10^{\pm3}$ according to cross-validation.
In SFS, we chose the value of $t_i=1$ or $t_i=-1$ to which class $i$ is set when using the number of classes as the number of reduced dimensions (Section~3.3), chose the number of reduced dimension $\ell$, and chose the value of $t_i$ again.
We used the KLFDA code from the Sugiyama--Sato--Honda Lab site\footnote{http://www.ms.k.u-tokyo.ac.jp/} and programmed the LADA and KDA codes.
All programs were coded and run in MATLAB 2018a. 

\subsection{Artificial datasets}
\begin{figure}[!h]
  \begin{center}
    \includegraphics[scale=0.4]{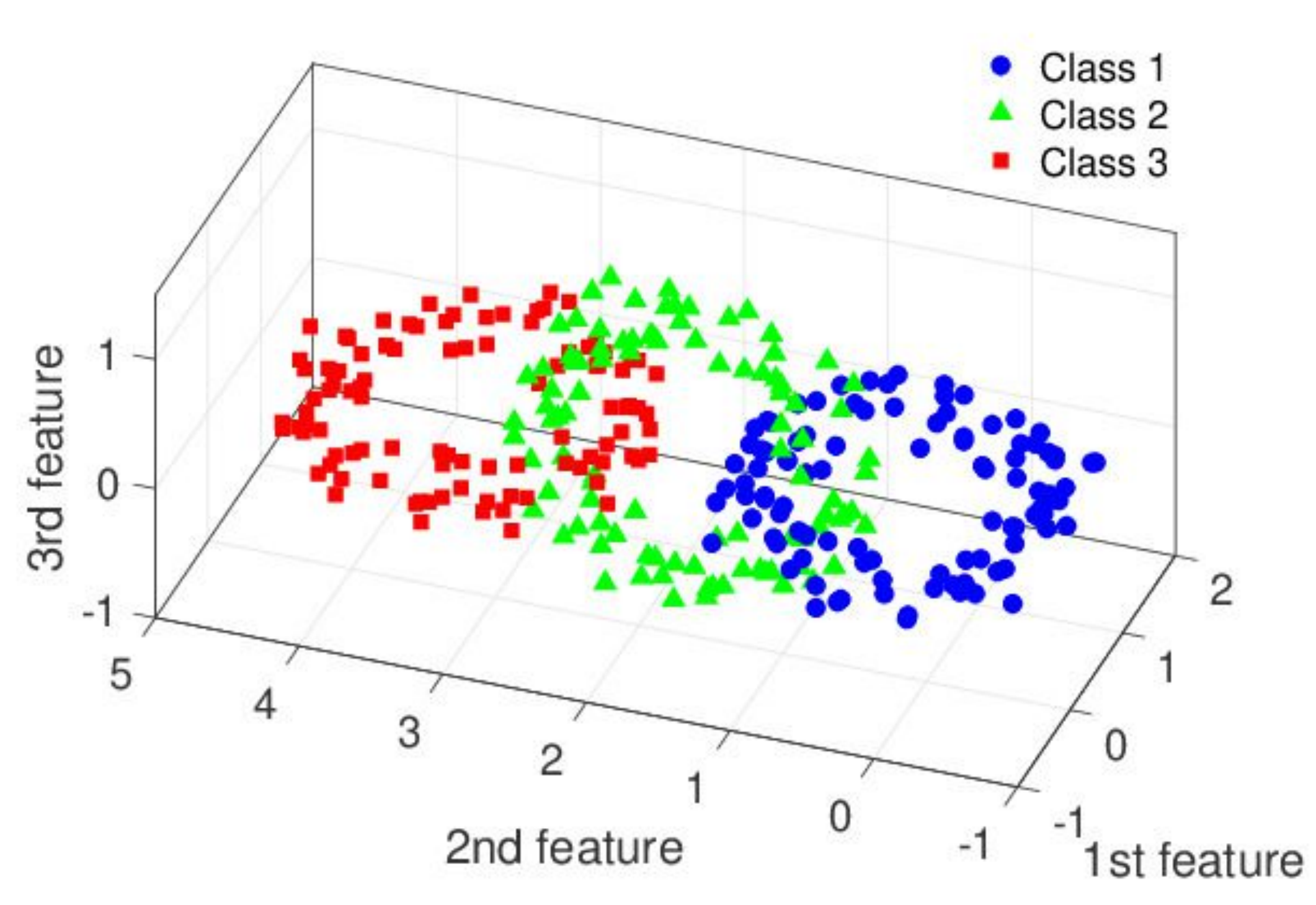}
    \caption{Visualization of the artificial data in three dimensions}
    \label{fig:arti}
  \end{center}

  \vspace{20pt}
  \begin{center}
    \includegraphics[width=1\hsize]{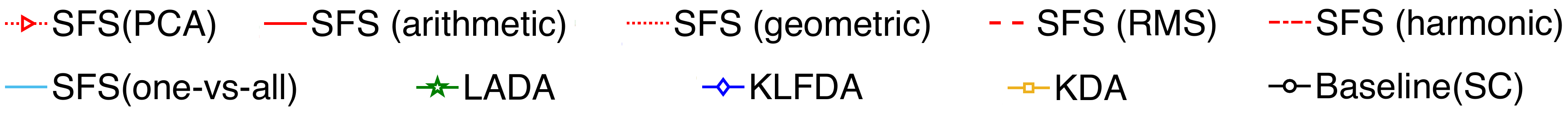}
  \end{center}
  \vspace{-10pt}
  \begin{minipage}[b]{0.33\hsize}
    \begin{center}
      \includegraphics[width=1\hsize]{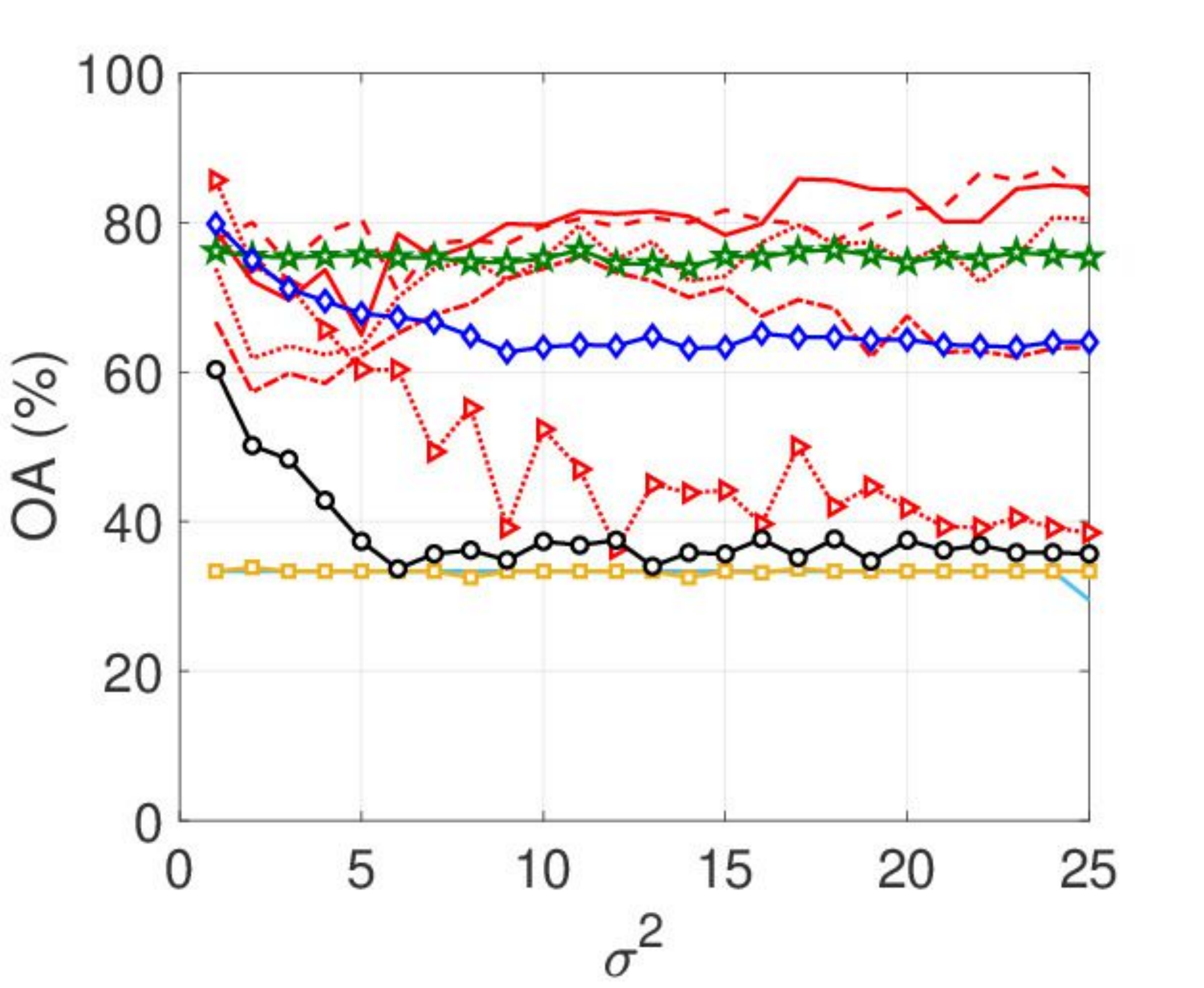}
    \end{center}
  \end{minipage}
  \begin{minipage}[b]{0.32\hsize}
    \begin{center}
      \includegraphics[width=1\hsize]{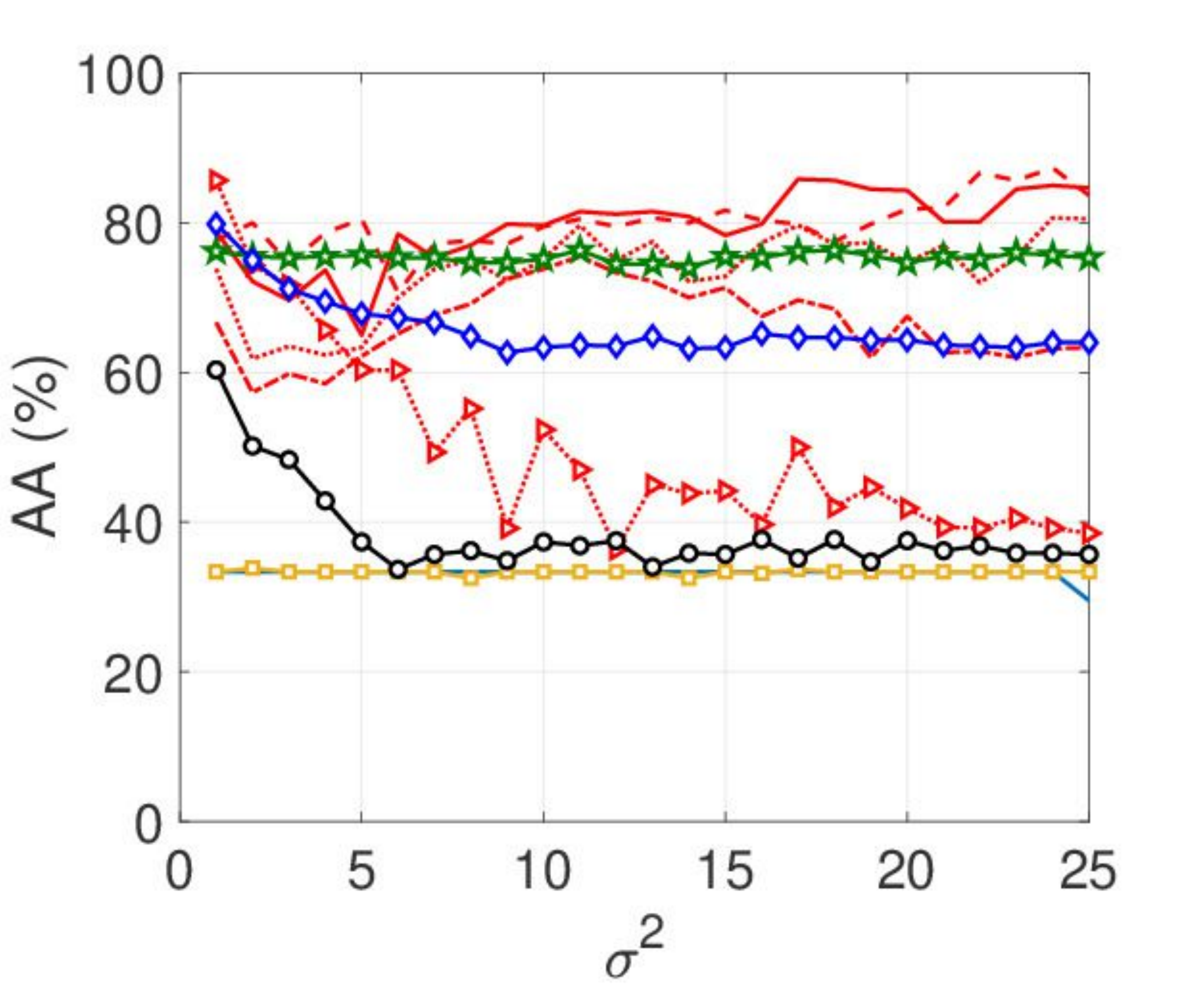}
    \end{center}
  \end{minipage}
  \begin{minipage}[b]{0.33\hsize}
    \begin{center}
      \includegraphics[width=1\hsize]{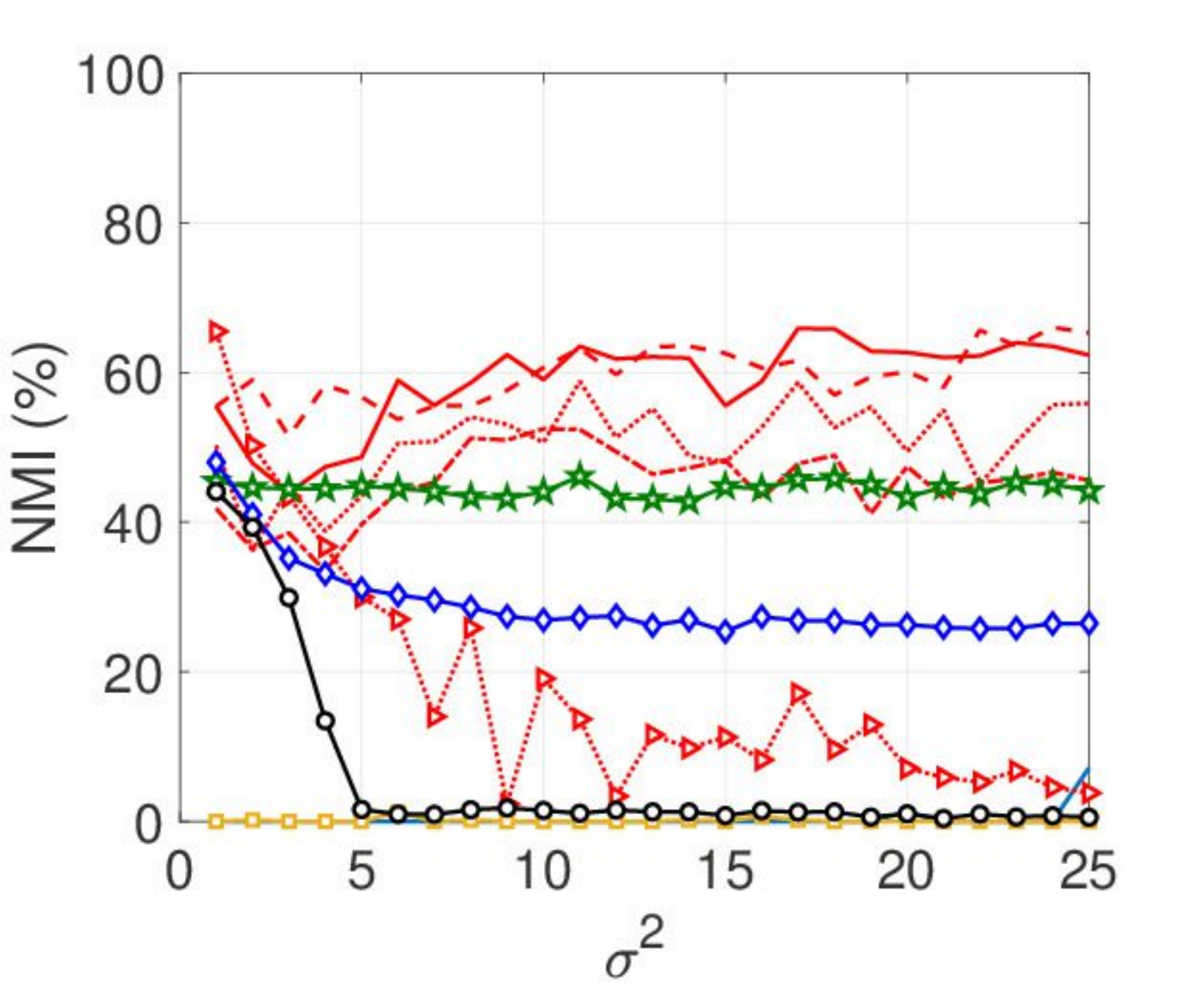}
    \end{center}
  \end{minipage}
  \caption{Mean of each performance measure vs.\ variance $\sigma^2$}
  \label{fig:accuracy}
\end{figure}

In this subsection, we report results on a toy problem consisting of 600 data samples with ten features.
The data samples had three classes, and each class had 200 samples.
Figure~\ref{fig:arti} shows three of the ten features, where the symbols $\circ$, $\vartriangle$, and $\square$ represent data samples in classes 1, 2, and 3, respectively, and the data samples in each class are normally distributed along a ring.
The rings of class 1 and 2, and those of classes 2 and 3 intersect, respectively.
The variances of the first to third features are 0.35, 2.01, and 0.19, respectively.
The remaining seven features are normally distributed with center zero and variance $\sigma^2$.
We changed the variance $\sigma^2$ from 1 to 25 in increments of 1.
These seven features interfere with data sample classification according to the first three features.
The classification of data samples with the reduced dimension was done by logistic regression.
Results for other classifiers are given in the Appendix.
In SFS, the number of prescribed eigenvectors $\bm{v}^{(p)}$ was set to the number of classes.
\par
Figure~\ref{fig:accuracy} shows the performance measures for variances $\sigma^2 = 1,2,\ldots, 25$ where SFS (one-versus-all) represents the ``one-vs-all'' approach~\cite[p.~658]{mSVM} for the previous SFS for a binary classification problem.
Regarding the performance measures, some of the proposed methods were more accurate than the previous methods, and KDA was the worst regardless of the variance.
SFS (PCA) was the best when the variance was small, while SFS(arithmetic) and SFS (RMS) were the best when the variance was large.
SFS(RMS), SFS(arithmetic), and LADA tended to be robust against interfering features.
SFS(one-vs-all) was almost as accurate as KDA, and the line was hidden by the KDA line.
LADA was more accurate than KLFDA, though LADA is not kernelized.
Among the compared SFS methods, the RMS approach was most accurate for many values of $\sigma^2$.
The proposed SFS methods were more accurate than the conventional ``one-vs-all'' approach for the multiclass classification problem.
\par
\begin{figure}[!h]
  \begin{minipage}{0.24\hsize}
    \begin{center}
      \includegraphics[width=1\hsize]{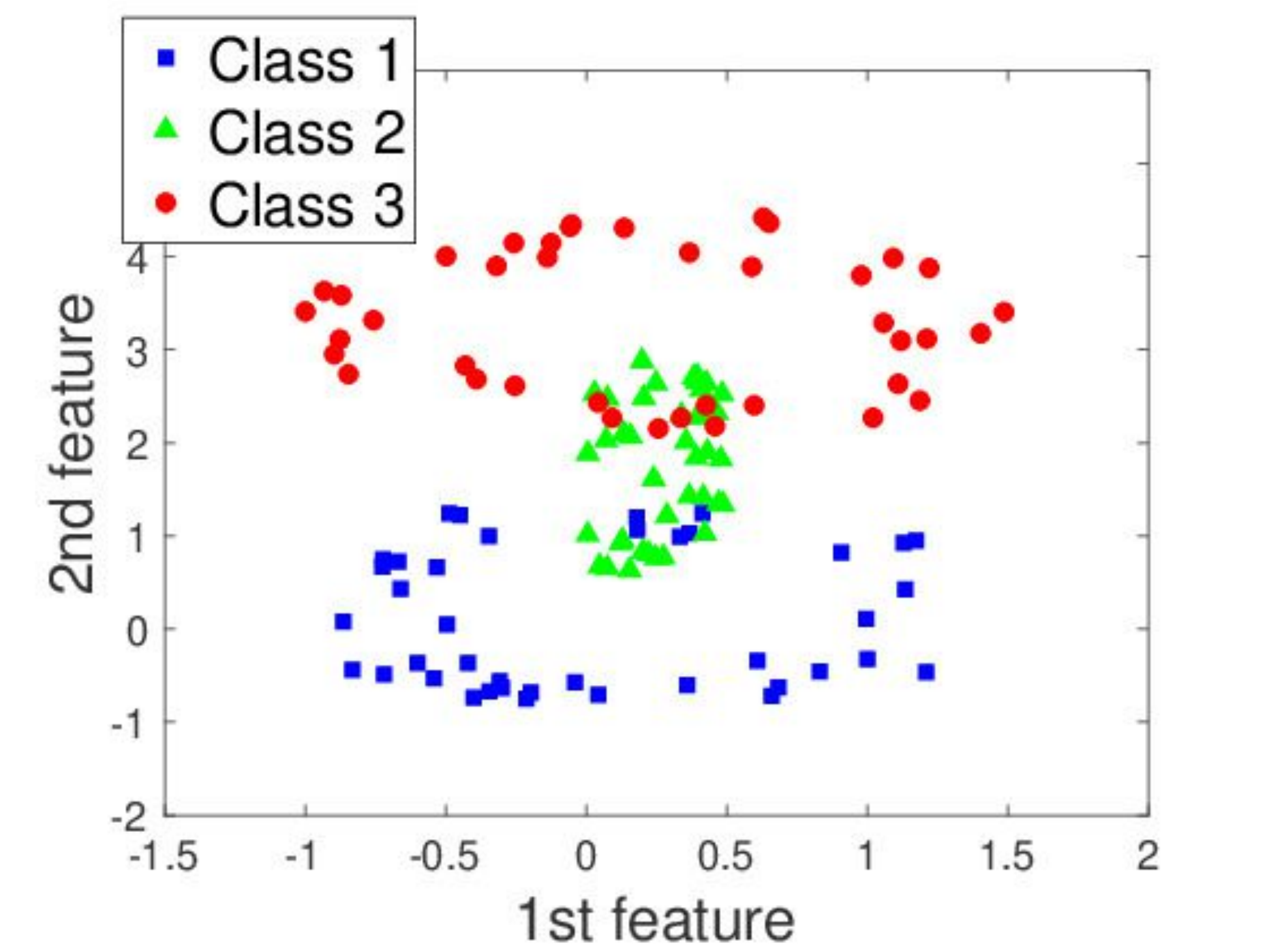}\\
      {\scriptsize (a) Original data}
    \end{center}
  \end{minipage}
  \hfill
  \begin{minipage}{0.24\hsize}
    \begin{center}
      \includegraphics[width=1\hsize]{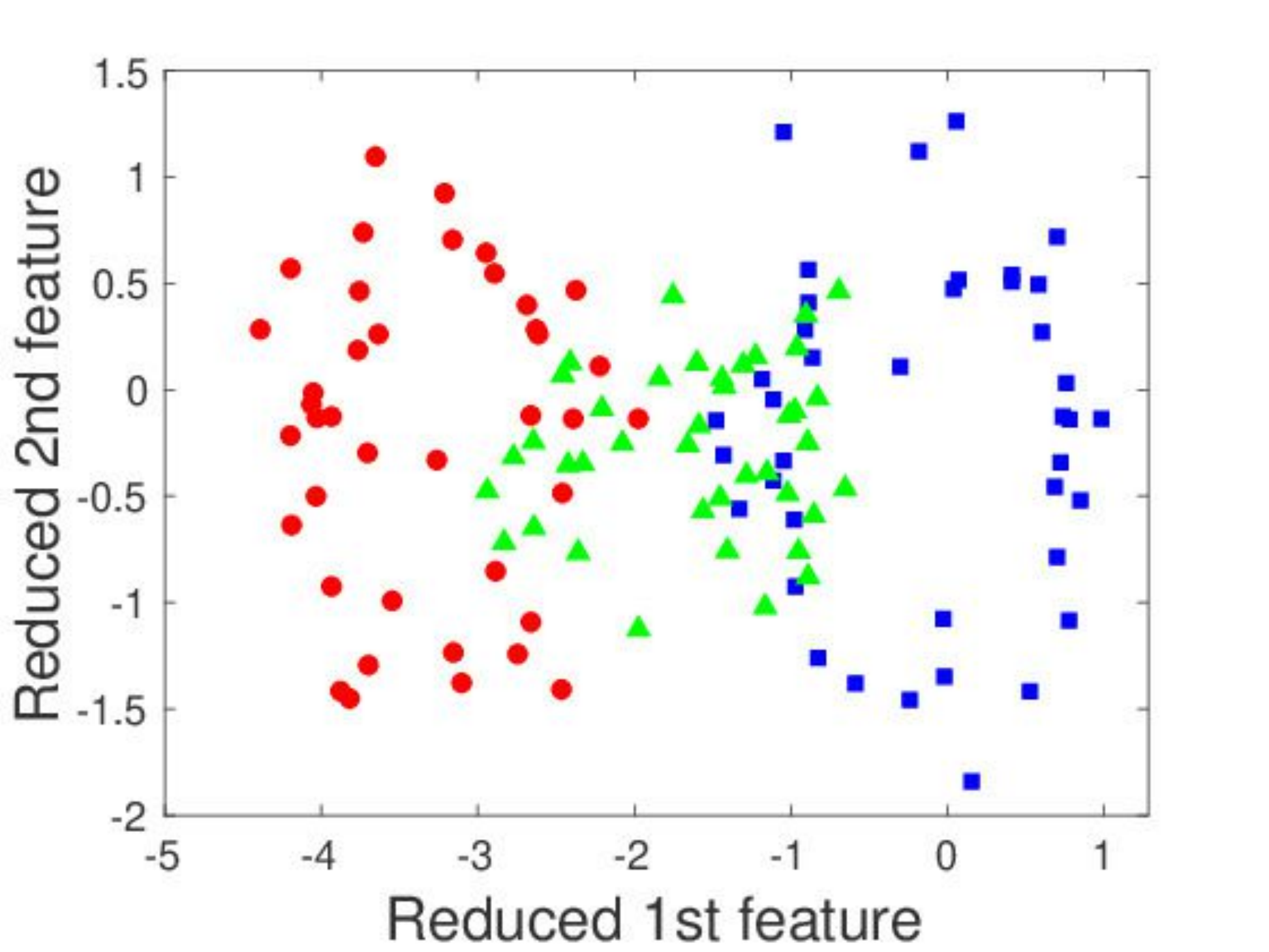}\\
      {\scriptsize (b) LADA}
    \end{center}
  \end{minipage}
  \hfill
  \begin{minipage}{0.24\hsize}
    \begin{center}
      \includegraphics[width=1\hsize]{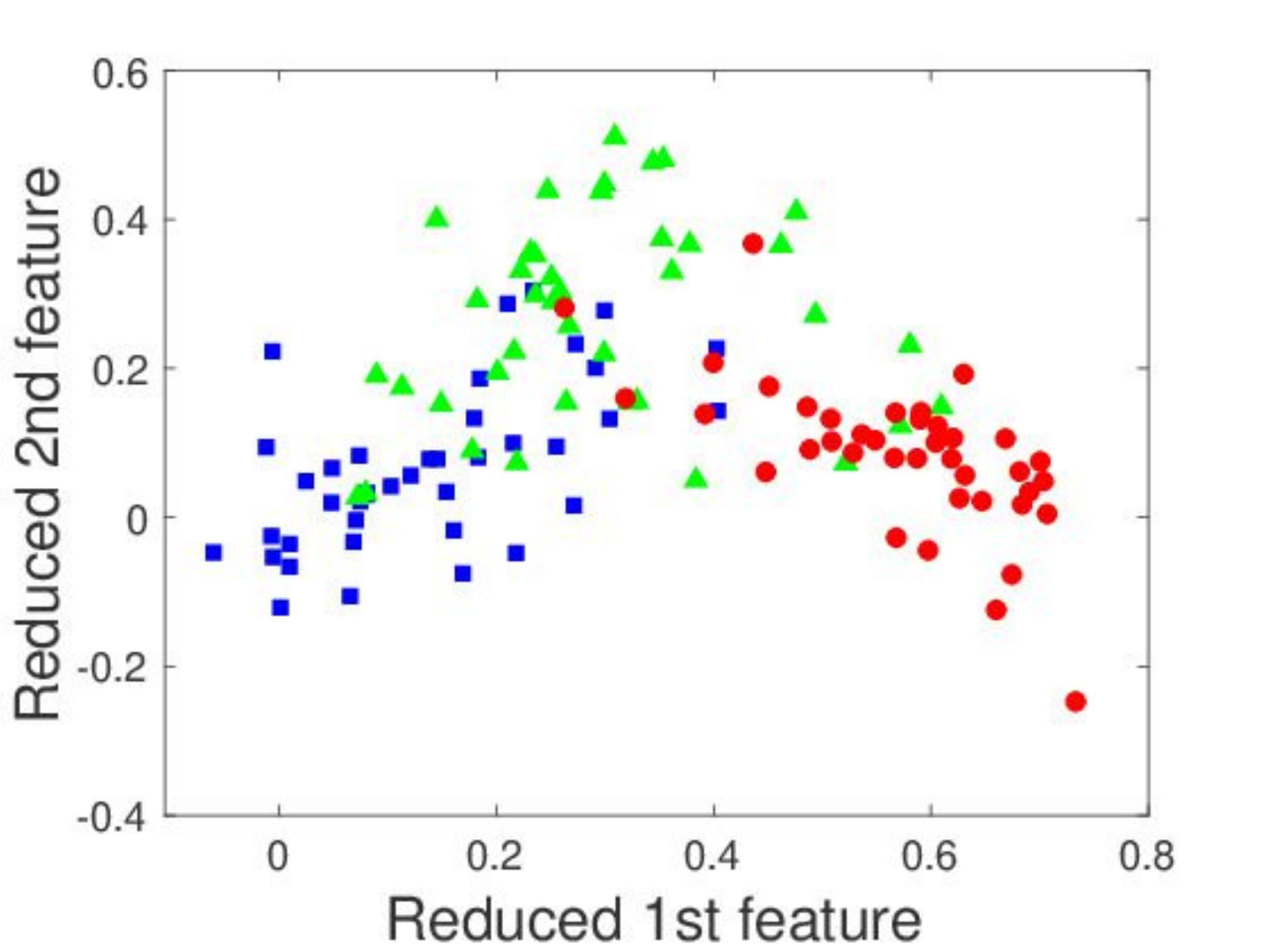}\\
      {\scriptsize (c) KLFDA}
    \end{center}
  \end{minipage}
  \hfill
  \begin{minipage}{0.24\hsize}
    \begin{center}
      \includegraphics[width=1\hsize]{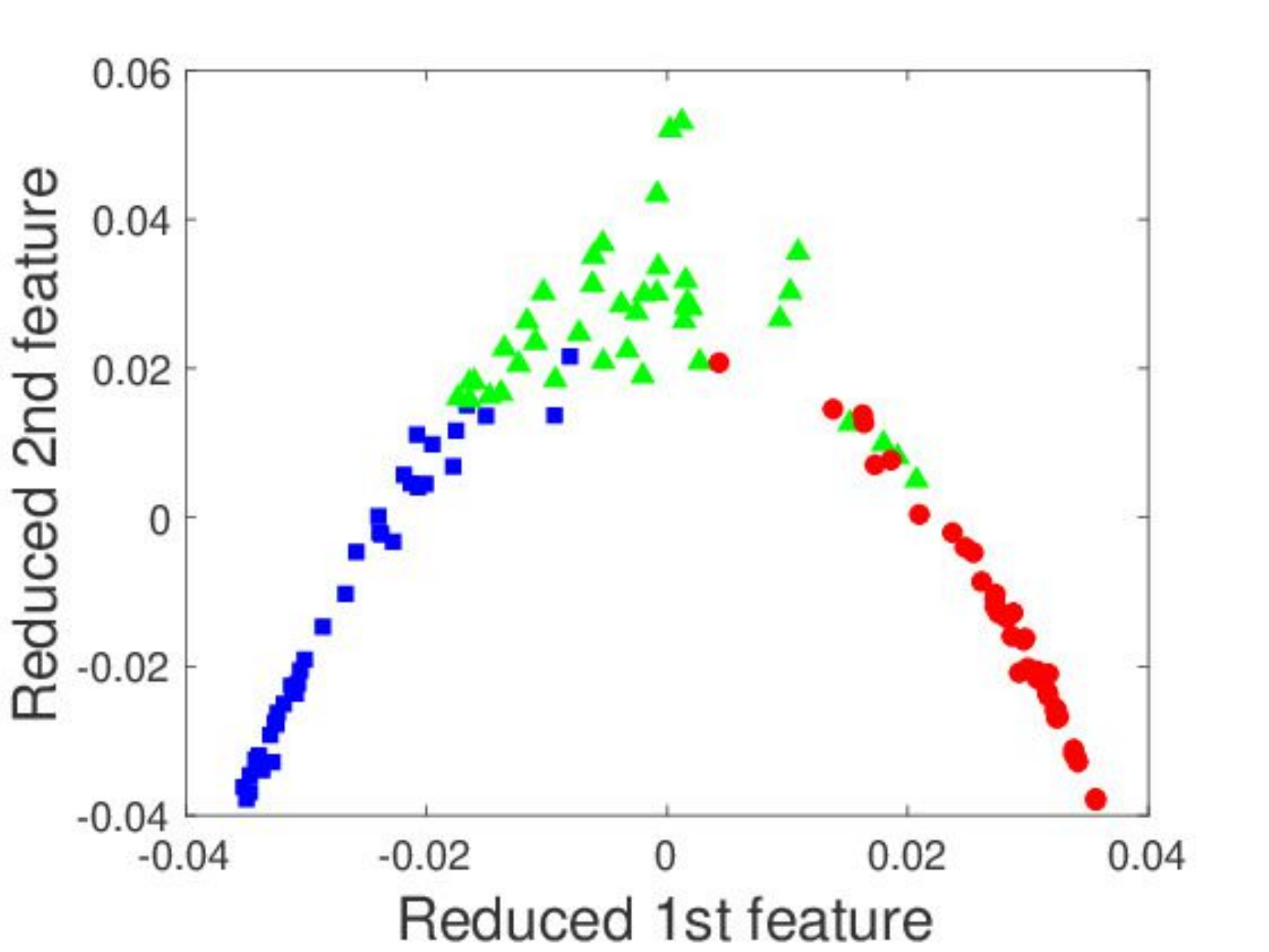}\\
      {\scriptsize (d) SFS (RMS)}\\
    \end{center}
  \end{minipage}
  \caption{Test data samples for $\sigma^2=1$ in the original and reduced two-dimensional space}
  \label{fig:2dim}
\end{figure}
\begin{figure}  
  \begin{minipage}{0.24\hsize}
    \vspace{20pt}
    \begin{center}
      \includegraphics[width=1\hsize]{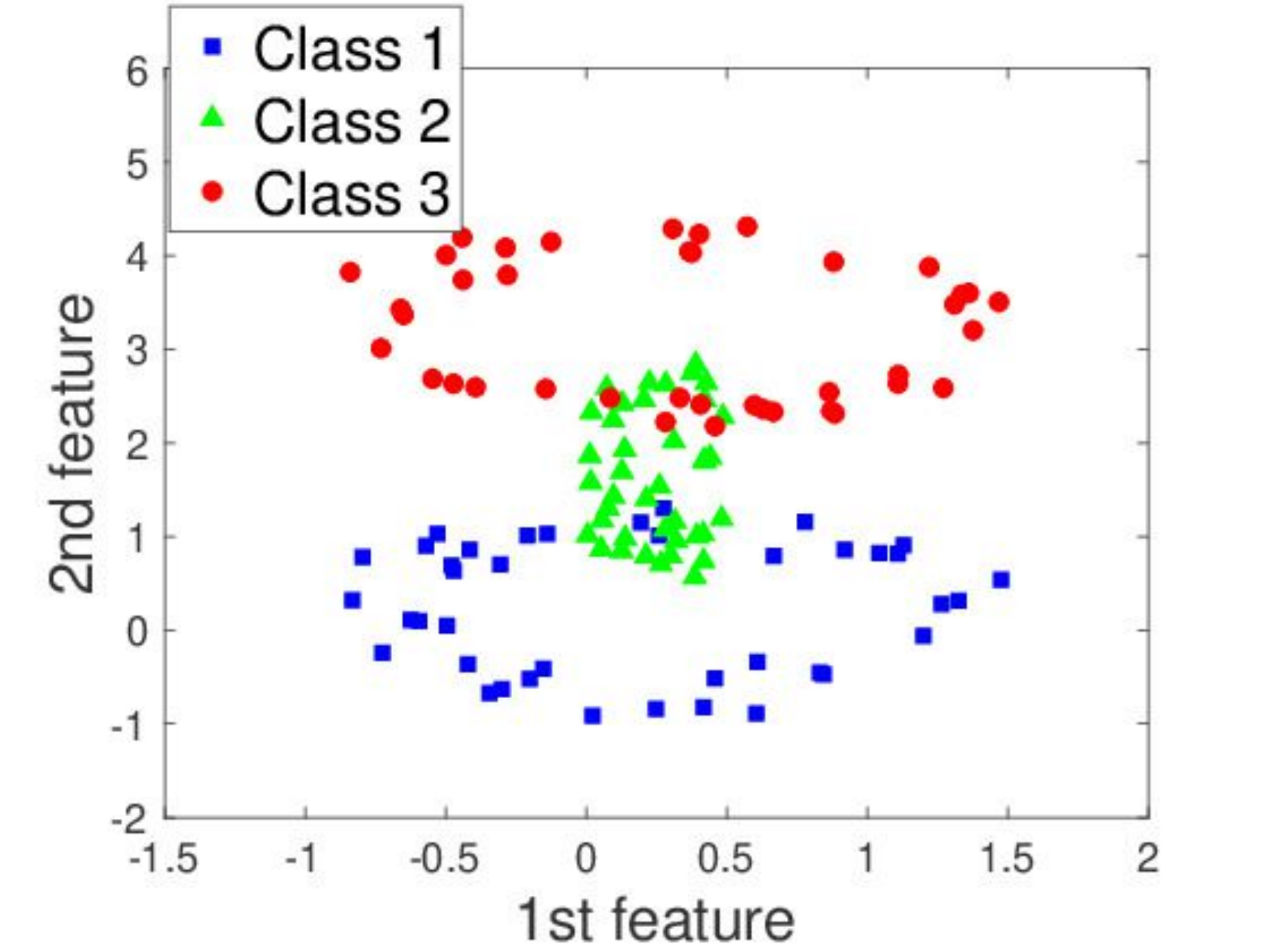}\\
      {\scriptsize (a) Original data}
    \end{center}
  \end{minipage}
  \hfill
  \begin{minipage}{0.24\hsize}
    \begin{center}
      \vspace{20pt}
      \includegraphics[width=1\hsize]{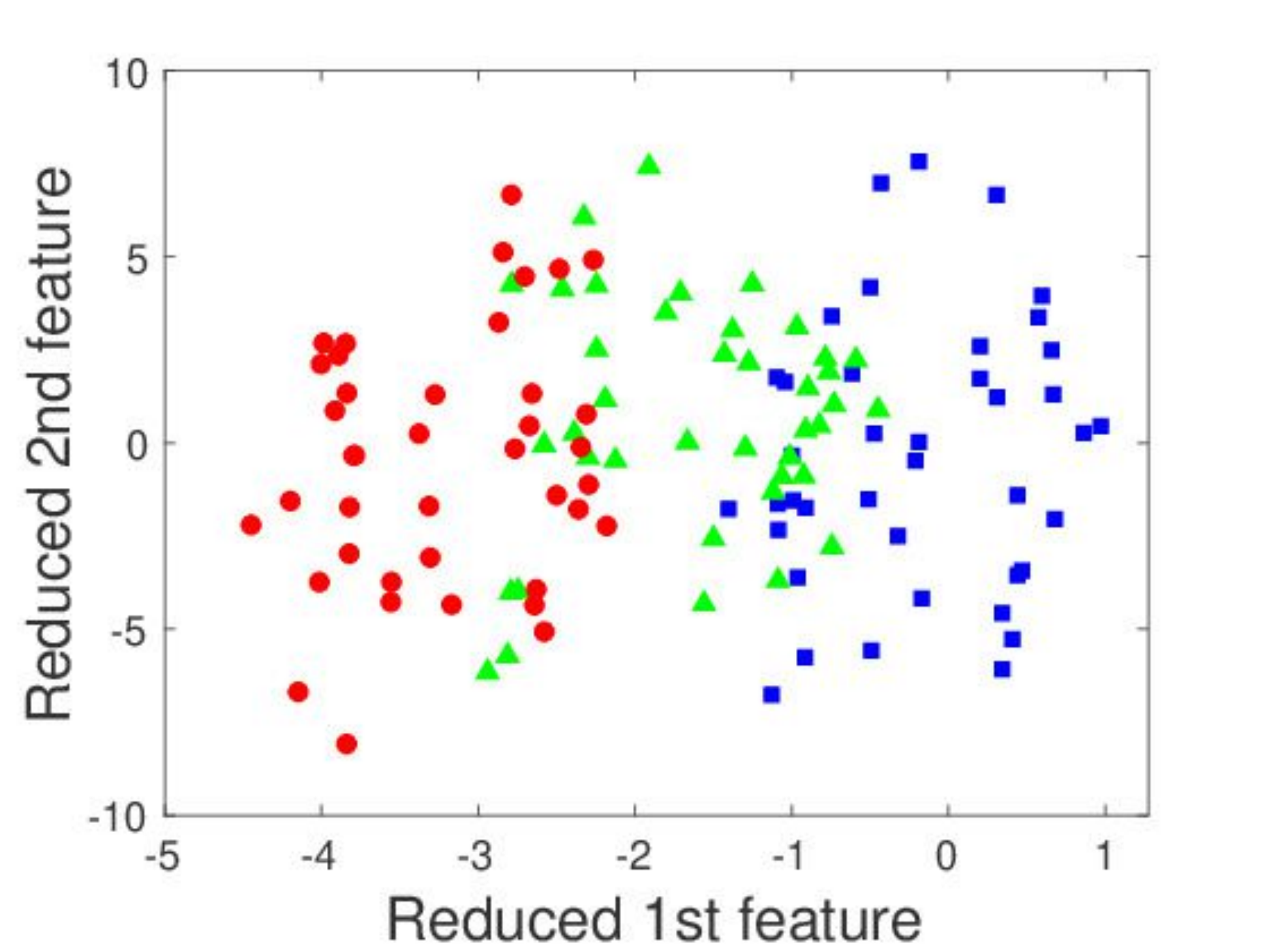}\\
      {\scriptsize (b) LADA}
    \end{center}
  \end{minipage}
  \hfill
  \begin{minipage}{0.24\hsize}
    \begin{center}
      \vspace{20pt}
      \includegraphics[width=1\hsize]{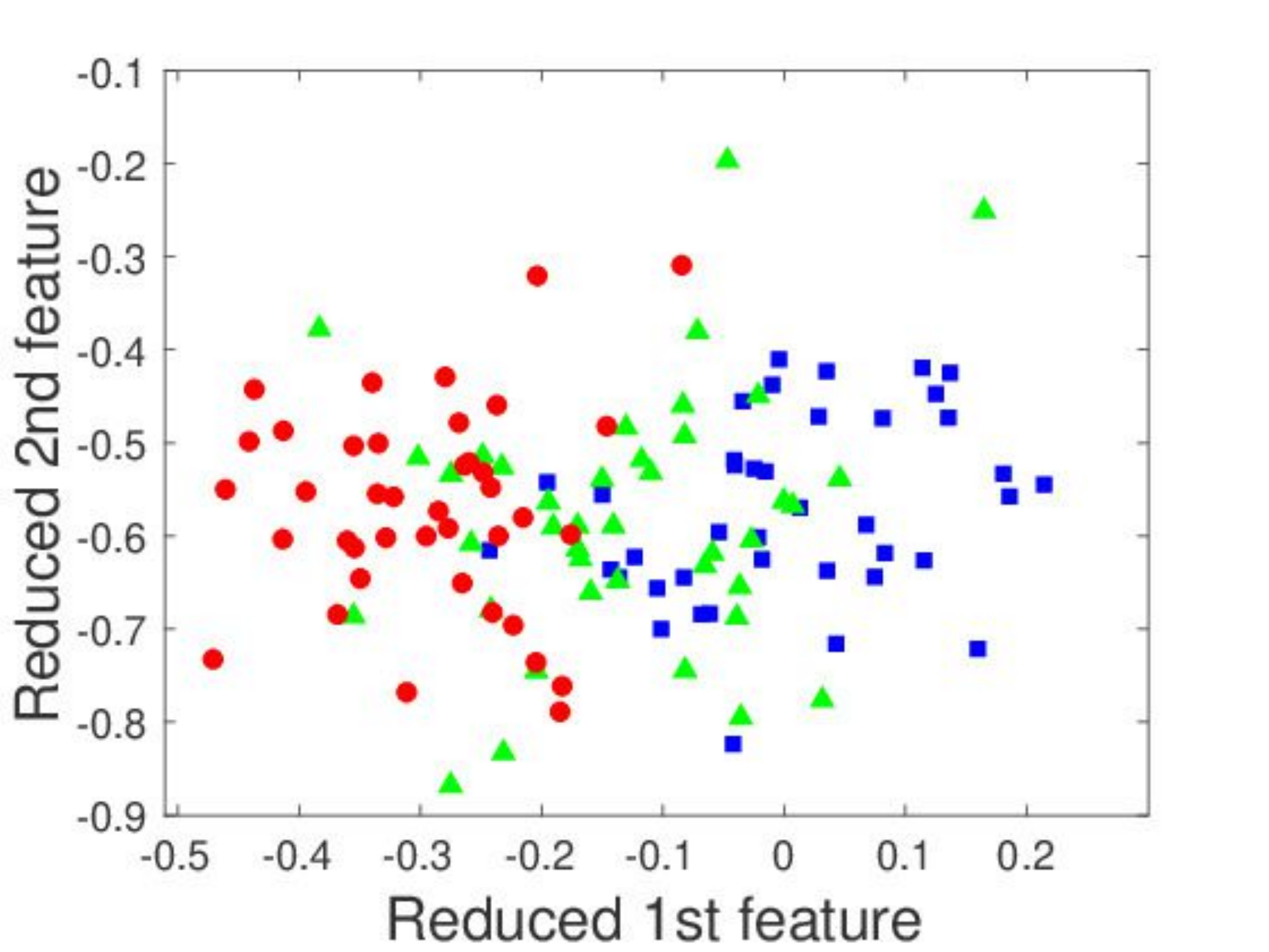}\\
      {\scriptsize (c) KLFDA}
    \end{center}
  \end{minipage}
  \hfill
  \begin{minipage}{0.24\hsize}
    \begin{center}
      \vspace{20pt}
      \includegraphics[width=1\hsize]{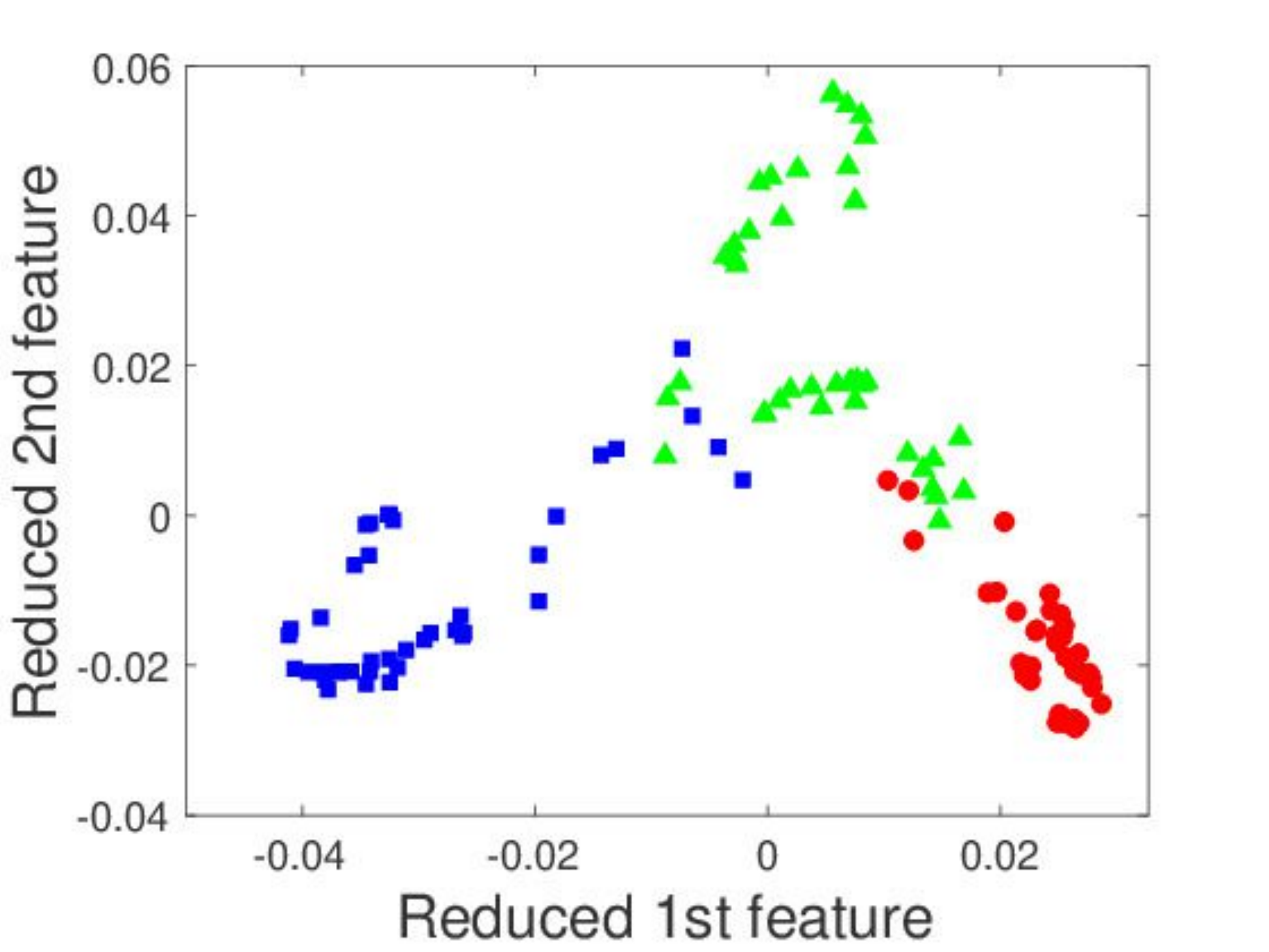}\\
      {\scriptsize (d) SFS (RMS)}
    \end{center}
  \end{minipage}
  \caption{Test data samples for $\sigma^2=25$ in the original and reduced two-dimensional space}
  \label{fig:2dim2}
\end{figure}

Figure~\ref{fig:2dim} and \ref{fig:2dim2} show the first and second features of the original data in parts (a) and the two-dimensional data samples reduced by each method in parts (b)--(d) for $\sigma^2=1$ and $25$, respectively.
LADA and KLFDA failed to enhance the separations of clusters, while the proposed method separated clusters well.
\par
\begin{figure}[!h]
  \begin{minipage}{0.5\hsize}
    \begin{center}
      \includegraphics[width=.9\hsize]{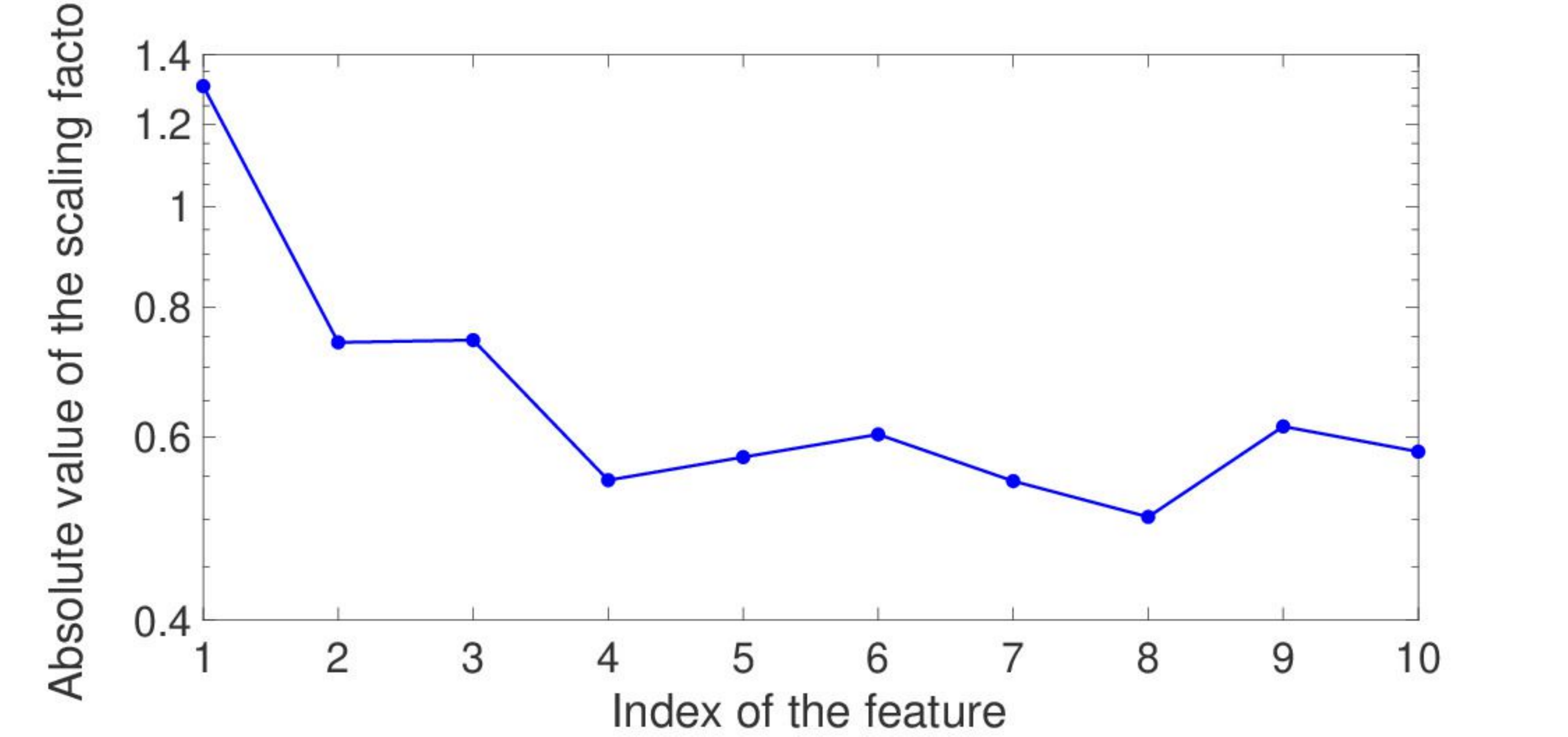}\\
      {\scriptsize (a) $\sigma^2 = 1$}
    \end{center}
  \end{minipage}
  \hfill
  \begin{minipage}{0.5\hsize}
    \begin{center}
      \includegraphics[width=.9\hsize]{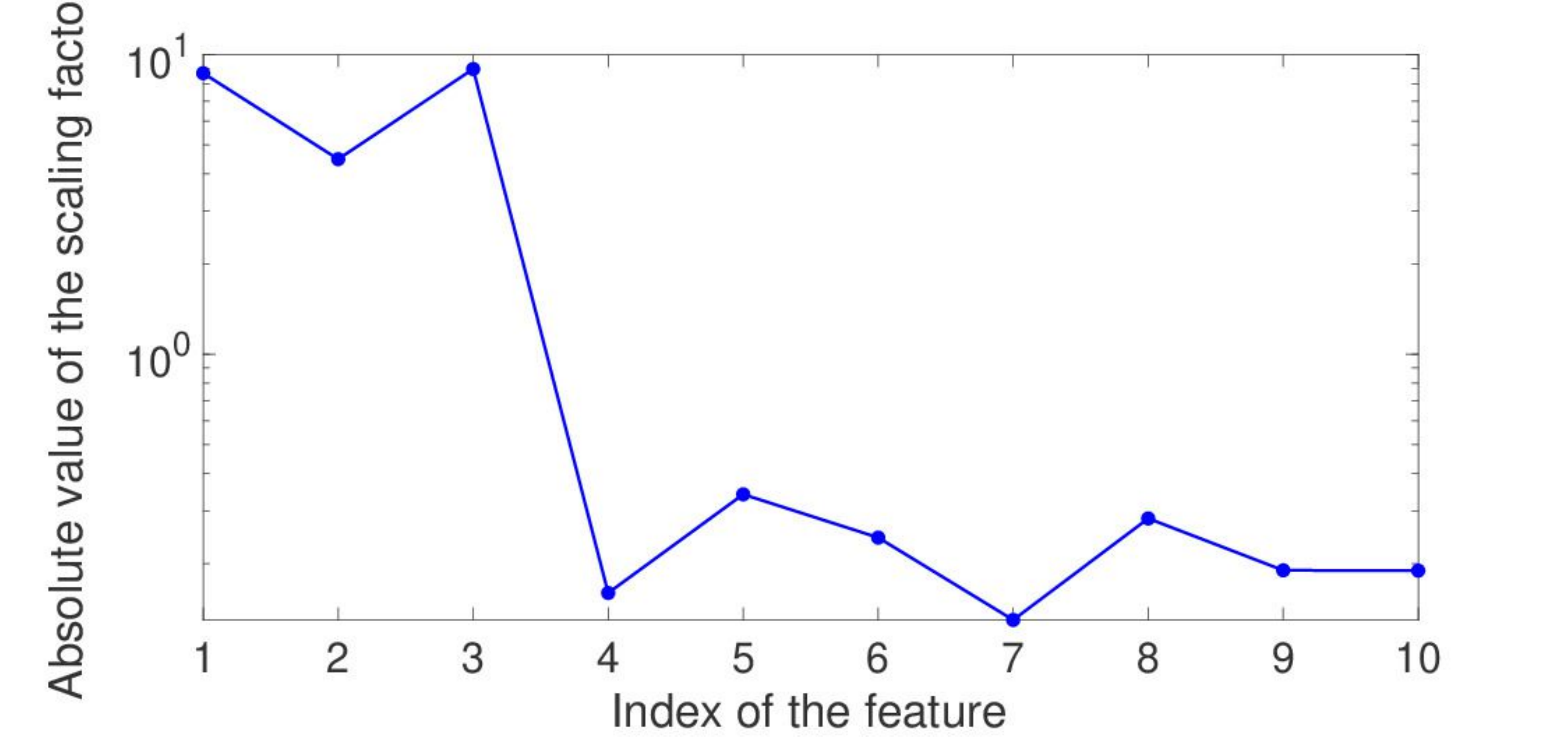}\\
      {\scriptsize (b) $\sigma^2 = 25$}
    \end{center}
  \end{minipage}
  \caption{Values of the scaling factors for (a) $\sigma^2=1$ and (b) $\sigma^2=25$ when using the root mean square}
  \label{fig:SF}
\end{figure}

Figure~\ref{fig:SF} shows the absolute values of the obtained scaling factors of SFS(RMS) for $\sigma^2=1$ and $\sigma^2=25$.
Because the fourth to tenth features given by random numbers are not involved in the classification, the values of the scaling factors of these seven features were small.
Thus, the values of the scaling factors indicate which features are effective in the desired classification.

\subsection{Real-world datasets}
In this section, we report results for five datasets with more features than samples in the real world from a public repository\footnote{http://featureselection.asu.edu/datasets.php}.
Table~\ref{tab:1} gives the numbers of samples, features, and classes of the datasets.
Datasets (a)--(d) are of gene expressions, and dataset (e) is of face images.
In SFS, the number of prescribed eigenvectors was set to the number of classes, and we used RMS to generate the feature scaling matrix.
The classification was done by KNN for $k=1$ since it was the best in the experiment in Section 5.1.

Table~\ref{tab:2} gives the mean and standard deviation of the performance measures for each method.
For the colon and lymphoma, the NMI of KDA was Not-a-Number (NaN) because KDA gave the same label to all test data samples.
In all performance measures, SFS (RMS) was the best or second best for all datasets.
KLFDA was the best for warpPIE10P.

\begin{table}[!h]
  \begin{center}
    \caption{Specifications of datasets}
    \label{tab:1}
    \begin{tabular}{l|ccc}
      & \# of samples & \# of features & \# of classes\\ \hline
      (a) Colon & 62 & 2000	& 2\\
      (b) CLL\_SUB\_111 & 111 & 11340 & 3 \\
      (c) Lung & 203 & 3312 & 5 \\
      (d) Lymphoma & 96 & 4026 & 9\\
      (e) WarpPIE10P & 210 & 2420 & 10
    \end{tabular}
  \end{center}
\end{table}

  \begin{table}[!h]
    \caption{Performance rates (mean\% $\pm$ std) for real-world datasets}
    \label{tab:2}
{\small
    \begin{center}
    (a) Colon\\ [5pt]
      \begin{tabular}{l|ccc}
        & OA & AA & NMI\\ \hline
        {\bf SFS (RMS)} & \underline{81.7$_{6.2}$} & {\bf 81.3$_{8.8}$} & \underline{35.1$_{19.2}$}\\
        LADA & 63.3$_{6.7}$ & 57.5$_{8.3}$  & 12.5$_{8.0}$\\
        KLFDA & {\bf 83.3$_{7.5}$} & \underline{78.8$_{8.5}$} & {\bf 36.7$_{22.2}$}\\
        KDA & 66.7$_{0.0}$ & 50.0$_{0.0}$ & NaN\\
      \end{tabular}\\ [15pt]
      
%
%
    \begin{center}
    (b) CLL\_SUB\_111\\[5pt]    
      \begin{tabular}{l|ccc}
        &  OA & AA & NMI\\ \hline
        {\bf SFS (RMS)} & {\bf 68.2$_{5.0}$} & {\bf 72.7$_{8.8}$} & {\bf 37.3$_{2.2}$}\\
        LADA & 53.6$_{7.3}$ & 38.0$_{6.5}$ & 11.2$_{8.3}$\\
        KLFDA & \underline{67.3$_{7.8}$} & \underline{70.7$_{9.3}$} & \underline{34.2$_{7.9}$}\\
        KDA & 50.0$_{5.0}$ & 35.3$_{4.0}$ & 15.9$_{13.3}$\\
      \end{tabular}\\ [15pt]
      
              (c) Lung\\ [5pt]
      \begin{tabular}{l|ccc}
        & OA & AA & NMI\\ \hline
        {\bf SFS (RMS)} & {\bf 94.5$_{1.9}$} & {\bf 85.4$_{11.1}$} & {\bf 85.7$_{6.3}$}\\
        LADA & \underline{93.5$_{3.4}$} & 79.4$_{13.1}$  & \underline{79.8$_{10.2}$}\\
        KLFDA & 93.0$_{2.9}$ & \underline{82.0$_{10.7}$} & 78.7$_{8.3}$\\
        KDA & 64.5$_{8.7}$ & 20.4$_{4.4}$ & 8.5$_{10.0}$\\
      \end{tabular}\\ [15pt]
            \end{center}

    (d) Lymphoma\\ [5pt]
        \begin{tabular}{l|ccc}
          & OA & AA & NMI\\ \hline
          {\bf SFS (RMS)} & {\bf 94.8$_{9.2}$} & {\bf 94.8$_{9.2}$} & {\bf 94.6$_{5.2}$}\\
          LADA & 60.0$_{7.8}$ & 42.8$_{15.4}$ & 62.7$_{7.4}$\\
          KLFDA & \underline{90.5$_{2.1}$} & \underline{80.0$_{4.4}$} & \underline{90.7$_{2.2}$}\\
          KDA & 47.4$_{0.0}$ & 11.1$_{0.0}$ & NaN\\
        \end{tabular}\\ [15pt]
      \end{center}
    
\begin{center}
	(e) WarpPIE10P\\ [5pt]
        \begin{tabular}{l|ccc}
          & OA & AA & NMI\\ \hline
          {\bf SFS (RMS)} & \underline{88.6$_{2.8}$} & \underline{88.9$_{2.6}$} & \underline{90.3$_{2.4}$}\\
          LADA & 82.9$_{2.8}$ & 82.6$_{3.2}$ & 83.7$_{3.0}$\\
          KLFDA & {\bf 99.5$_{1.0}$} & {\bf 99.5$_{1.0}$} & {\bf 99.5$_{1.0}$}\\
          KDA & 23.3$_{7.6}$ & 8.7$_{2.5}$ & 24.1$_{14.1}$
        \end{tabular}
        \end{center}
        }
\end{table}

\section{Conclusions}
In this paper, to deal with irregularity or uncertainty in features, we extended previously proposed feature scaling methods to multiclass classification and proposed a supervised dimensionality reduction method that exploits knowledge of labels of training data samples.
The proposed method aggressively modifies the scales of features, and feature scaling can reduce the effects of features that prevent us from obtaining the desired clusters.
To obtain the factors for scaling the features based on spectral clustering, we derived matrix eigenproblems whose eigenvectors have the feature scaling factors and described the procedures for the proposed method.
The multiclass extension asks for more hyperparameters and computational cost.
To reduce the number of hyperparameters, we proposed an automatic tuning technique based on the formula for discrete optimization to determine the entries of prescribed eigenvectors.
The demanding combinations of entries of prescribed vectors are related to the number of eigenproblems to solve and can be reduced with a trade-off between accuracy and computational cost.
Numerical experiments showed that feature scaling is effective when combined with the proposed classification method.
For toy problems with more samples than features, feature scaling improved accuracy and was more robust to interfering features.
In terms of flexibility in merging candidates of scaling factors for each feature, our experimental results indicate that the RMS approach is preferable in practice.
For real-world problems with more features than samples, the proposed method was more robust than previous methods and outperformed previous methods in some cases. 

\section*{Acknowledgements}
The second author was supported in part by JSPS KAKENHI Grant Number 16K17639 and the Hattori Hokokai Foundation. The third author was supported in part by the Japan Science and Technology Agency, ACT-I (No.~JPMJPR16U6), New Energy and Industrial Technology Development Organization, and JSPS KAKENHI Grant Numbers 17K12690, 18H03250.

\appendix
\def\thesection{Appendix \Alph{section}.}
\section{Comparison of classifiers}
In Section 5.1, we used logistic regression to classify data samples with reduced dimensions.
In fact, the proposed method is not specialized to a particular classifier, which means we can choose an arbitrary classifier after the dimensionality reduction.
In this appendix, we observe the dependence of the accuracy on classifiers and compare four classifiers: logistic regression (LR), $k$-nearest neighbor algorithm (KNN), support vector machine (SVM), and naive Bayes classifier (NB).
Among the proposed methods, we used SFS (RMS) because it performed the best in many cases.

\begin{figure}[h]
	\begin{minipage}{0.47\hsize}
		\begin{center}
		\includegraphics[width=1\hsize]{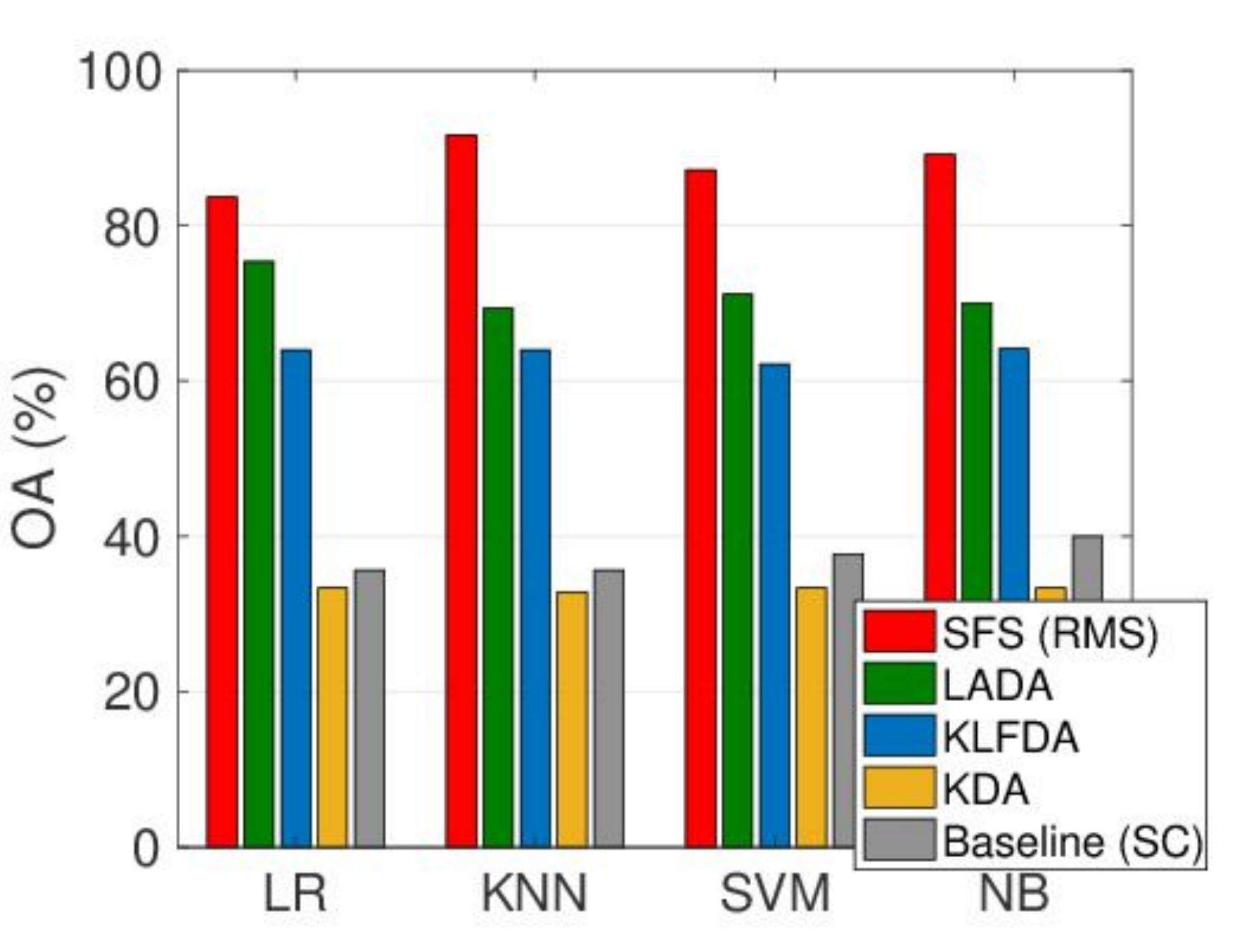}
		\caption{Mean performance measure (OA) vs.\ classifier}
		\label{fig:classifier}
		\end{center}
	\end{minipage}
	\hfill
	\begin{minipage}{0.47\hsize}
		\begin{center}
		\includegraphics[width=1\hsize]{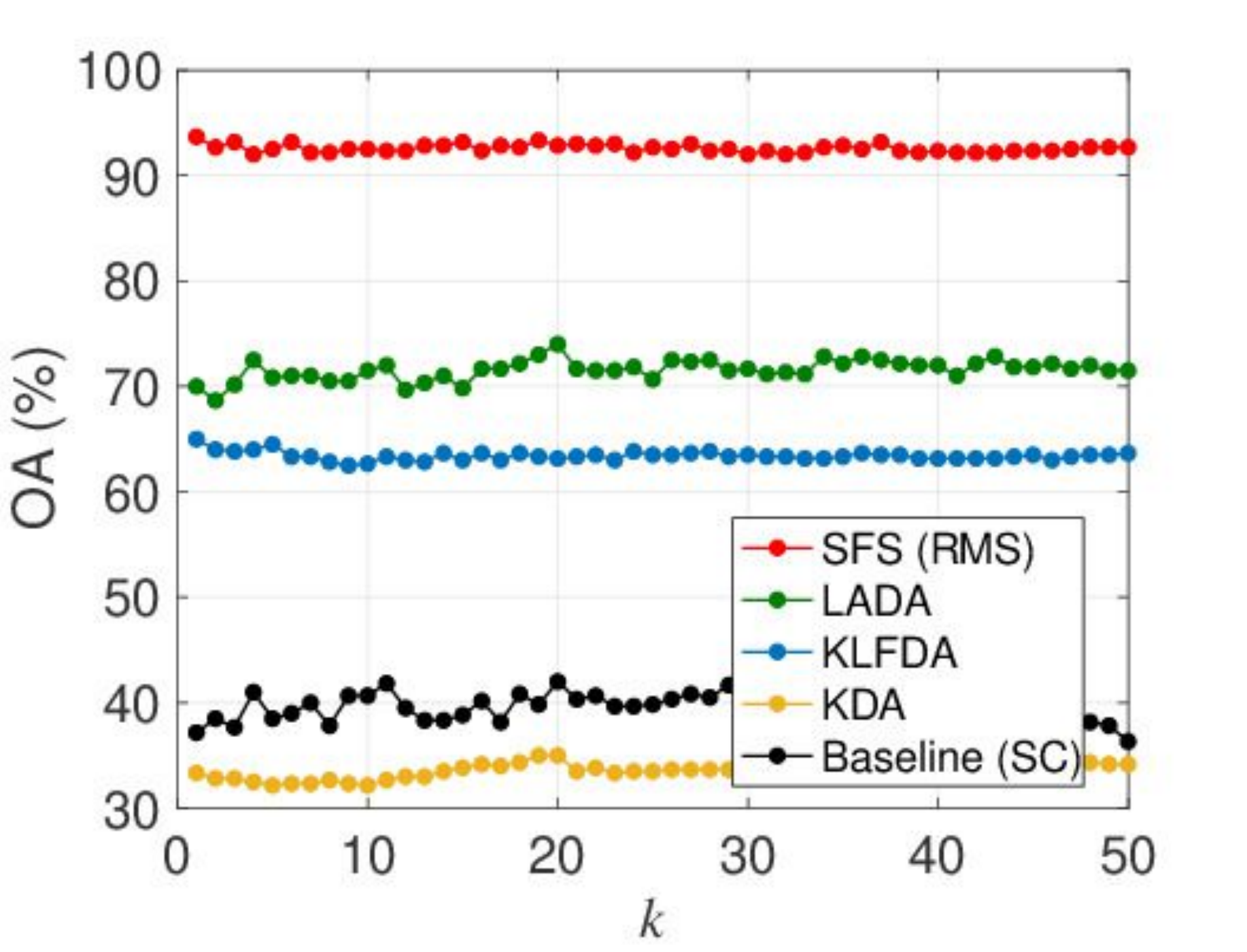}
		\caption{Mean of performance measure (OA) vs.\  $k$ in the $k$-NN classifier}
		\label{fig:KNN}
		\end{center}
	\end{minipage}
\end{figure}

Figure~\ref{fig:classifier} shows the mean OA for each method and each classifier.
SFS(RMS) was more accurate than other methods for every classifier.
SFS (RMS) with KNN was more accurate than the other classifiers.
The relative positions of the compared methods are the same for all performance measures.

Figure~\ref{fig:KNN} shows the OA of each method for $k=1,2,\ldots,50$ in the KNN.
For $k$=1, SFS (RMS) was the best and its OA was 93.7\%.
The accuracies of the compared methods did not largely depend on the value of $k$.
The relative positions of the compared methods are also the same for all $k$ in the KNN.

\bibliographystyle{abbrv}
\bibliography{multi}

\end{document}